\documentclass{article}

\PassOptionsToPackage{round}{natbib}

\usepackage[preprint]{neurips_2021}

\usepackage[utf8]{inputenc} 
\usepackage[T1]{fontenc}    
\usepackage{hyperref}       
\usepackage{url}            
\usepackage{booktabs}       
\usepackage{amsfonts}       
\usepackage{nicefrac}       
\usepackage{microtype}      
\usepackage{xcolor}         

\usepackage{import}
\usepackage{preamble}

\title{Investigating Alternatives to the Root Mean Square for Adaptive Gradient Methods}

\author{%
  Brett Daley\\
  Khoury College of Computer Sciences\\
  Northeastern University\\
  Boston, MA 02115 \\
  \texttt{b.daley@northeastern.edu}
  \And
  Christopher Amato\\
  Khoury College of Computer Sciences\\
  Northeastern University\\
  Boston, MA 02115\\
  \texttt{c.amato@northeastern.edu}
}

\begin{document}

\maketitle

\begin{abstract}
    Adam is an adaptive gradient method that has experienced widespread adoption due to its fast and reliable training performance.
    Recent approaches have not offered significant improvement over Adam, often because they do not innovate upon one of its core features:
    normalization by the root mean square (RMS) of recent gradients.
    However, as noted by~\cite{kingma2015adam}, any number of $L^p$ normalizations are possible, with the RMS corresponding to the specific case of $p=2$.
    In our work, we theoretically and empirically characterize the influence of different $L^p$ norms on adaptive gradient methods for the first time.
    We show mathematically how the choice of $p$ influences the size of the steps taken, while leaving other desirable properties unaffected.
    We evaluate Adam with various $L^p$ norms on a suite of deep learning benchmarks, and find that $p > 2$ consistently leads to improved learning speed and final performance.
    The choices of $p=3$ or $p=6$ also match or outperform state-of-the-art methods in all of our experiments.
\end{abstract}

\tightsection{Introduction}

Adaptive gradient methods have been a significant boon to deep learning by enabling fast training of high-dimensional networks while largely circumventing costly hyperparameter searches.
Deep learning models now routinely have millions of parameters to be optimized, and state-of-the-art architectures for language can exceed billions~\cite[e.g.][]{jozefowicz2016exploring, radford2019language, shoeybi2019megatron, raffel2019exploring, brown2020language}.
Such models are too large for Hessian-based optimization to be feasible, making fast first-order methods imperative for reducing training time.
Adaptive gradient methods achieve this by adjusting a separate learning rate for each parameter, often facilitating faster learning than Stochastic Gradient Descent (SGD) can.

Adam~\citep{kingma2015adam} is one adaptive gradient method that has been extraordinarily successful.
The method has garnered a reputation for reliability due to its fast empirical performance across a range of network architectures, even when using its default hyperparameters.
Because of Adam's strong performance, recent state-of-the-art methods have not deviated significantly from its formulation.
A non-exhaustive list of these include AMSGrad~\citep{reddi2019convergence}, Yogi~\citep{zaheer2018adaptive}, AdaBound~\citep{luo2019adaptive}, and AdaBelief~\citep{zhuang2020adabelief}.
These methods closely resemble Adam, applying small adjustments in an effort to coax better performance from the algorithm.
Unfortunately, their similarity to Adam renders their benefits somewhat incremental or limited to specific cases, and they have not achieved Adam's level of popularity.

What makes these methods so similar to Adam is that they all rely on momentum~\citep{polyak1964some} that is element-wise normalized by the root mean square (RMS) of recent gradients.
RMS normalization has remained the dominant paradigm for adaptive gradient methods since AdaGrad~\citep{duchi2011adaptive}, ADADELTA~\citep{zeiler2012adadelta}, and RMSProp~\citep{tieleman2012lecture}.
In order to see continued improvement in optimization for deep neural networks, it may be fruitful to expand our search beyond algorithms that rely on this type of normalization.

In pursuit of this, we revisit a generalization of Adam based on the $L^p$ norm that was introduced by~\cite{kingma2015adam}.
Observing that the RMS of values $x_1, \ldots, x_n$ is proportional to
$\smash{ \nroot{p}{\sum_{i=1}^n \abs{x_i}^p} }$
for $p=2$, then an infinite spectrum of new normalization possibilities becomes available by considering the cases where $p \neq 2$.
The authors derived the AdaMax algorithm from this generalization by taking the limit as $p \to \infty$;
however, no analysis of intermediate $p$-values was conducted in the paper, as well as any following works to our knowledge.
As a result, it is presently unclear what properties these norms have and whether they would perform well in practice.

As such, the main objective of this work is to understand and characterize the implications of incorporating different $L^p$ norms into Adam---and, by extension, other methods---in order to motivate alternatives to RMS normalization.
We begin by justifying the $L^p$ norm by discussing Adam's relationship to AdaGrad and preconditioned SGD.
We then study to what extent Adam's properties are modified by the choice of $p$, finding that these are largely unchanged except for the step magnitude (and direction as a consequence).
To empirically test these effects, we evaluate Adam with different $L^p$ norms on a variety of challenging deep learning tasks.
Our results corroborate our theoretical predictions and demonstrate that norms where $p>2$ can be leveraged to obtain better final performance.
We conclude that the common $L^2$ norm is not necessarily the best choice in practice, suggesting avenues for improving adaptive gradient methods by exploring alternatives to RMS normalization.

\tightsection{Background}
\label{sect:background}

In this section, we explore the motivation behind Adam's use of RMS normalization by analyzing its relationship to two important predecessor methods:
AdaGrad and RMSProp.
Note that all arithmetic operations between vectors should be assumed to be element wise in our work.
For example, element-wise multiplication---not the dot product---is denoted by $x \cdot y$, and therefore $\smash{ x \cdot x = x^2 }$.

Our work considers a stochastic objective function $f \colon \mathbb{R}^d \mapsto \mathbb{R}$ that must be minimized in expectation with respect to its parameters $\smash{ \theta \in \mathbb{R}^d }$.
At each iterate $t$ of the optimization, the function emits a random value $f_t(\theta)$ whose gradient $g_t = \grad_\theta f_t(\theta)$ can be efficiently computed.
Direct computation of $\grad f(\theta)$ is assumed to be infeasible, as well as any higher-order derivatives of $f_t(\theta)$.
We do not assume that $f(\theta)$ is convex nor that each $g_t$ is identically distributed (i.e.\ stationarity).

In this framework, the simplest strategy for minimizing $f(\theta)$ is SGD, wherein the parameters are updated in proportion to the negative gradient:
$\theta_t \gets \theta_{t-1} - \alpha \cdot g_t$.
The constant of proportionality $\alpha > 0$ is known as the learning rate.
With appropriate annealing of $\alpha$ over time, SGD converges to a first-order stationary point~\citep{robbins1951stochastic}.

The main disadvantage of SGD is that the rate of convergence can be very slow when the function is ill conditioned~\citep{boyd2004convex}.
Faster convergence can be achieved by adjusting a learning rate for each dimension separately.
AdaGrad~\citep{duchi2011adaptive} ignited recent interest in these so-called \textit{adaptive gradient methods} for deep learning.
In its diagonal form, AdaGrad normalizes each gradient step by the square root of the sum of squared gradients:
\begin{equation}
    \label{eq:adagrad}
    - \alpha \cdot \frac{g_t}{\sqrt{\sum_{i=1}^t g_i^2}}
    \quad \text{(AdaGrad)}
\end{equation}
This formulation can be understood as an instantiation of preconditioned SGD, where the gradient is left-multiplied by a diagonal matrix
$\smash{ G_t^{\slashdiv{-1}{2}} }$ with $G_t \gets G_{t-1} + \diag(g_t^2)$.
While AdaGrad is effective for sparse-gradient problems, its empirical performance can degrade when gradients are dense or the objective function is nonconvex, since the component-wise learning rates are monotone non-increasing~\citep{goodfellow2016deep}.
RMSProp~\citep{tieleman2012lecture} resolves this issue by replacing the summation with an exponential moving average (EMA) with $\beta_2 \in [0,1)$:
\begin{equation}
    \label{eq:rmsprop}
    - \alpha \cdot \frac{g_t}{\sqrt{v_t}}
    \triangleq - \alpha \cdot \frac{g_t}{\sqrt{(1-\beta_2) \sum_{i=1}^t \beta_2^{t-i} \cdot g_i^2}}
    \quad \text{(RMSProp)}
\end{equation}
Comparing (\ref{eq:rmsprop}) with (\ref{eq:adagrad}), we can see that RMSProp is essentially a ``forgetful'' version of AdaGrad due to its EMA in the denominator.
RMSProp therefore behaves approximately like AdaGrad when $\beta_2 \approx 1$, but gradients arbitrarily far in the past have an arbitrarily small effect.
Note that the denominator $\sqrt{v_t}$ can now be interpreted as the root mean square (RMS) with exponential weights, but its motivation stems from AdaGrad's diagonal preconditioner, rather than the mathematical interpretation of the RMS itself.

Adam~\citep{kingma2015adam} is closely related to RMSProp, principally adding momentum~\citep{polyak1964some} to its formula:
\begin{equation}
    \label{eq:adam}
    - \alpha \cdot \frac{m_t}{\sqrt{v_t}}
    \triangleq - \alpha \cdot \frac{(1-\beta_1) \sum_{i=1}^t \beta_1^{t-i} \cdot g_i}{\sqrt{(1-\beta_2) \sum_{i=1}^t \beta_2^{t-i} \cdot g_i^2}}
    \quad \text{(Adam)}
\end{equation}
\cite{kingma2015adam} also introduce initialization bias correction, which we do not show here for brevity.
Although these changes appear to be relatively minor, they significantly improve empirical performance.
In particular, the momentum term $m_t$ interacts with the denominator $\sqrt{v_t}$ to regulate Adam's step sizes in a favorable way;
we discuss this in Section~\ref{sect:analysis}.

To summarize, AdaGrad utilizes a diagonal preconditioner based on the sum of the gradients' outer products, which RMSProp substitutes with an EMA to address nonconvexity and nonstationarity.
Adam then adds momentum and bias correction to RMSProp.
The view of the exponentially weighted RMS as a moving approximation to AdaGrad's diagonal preconditioner is important;
it suggests that the specific statistical interpretation of the RMS itself is not critical, and therefore it should be possible to substitute other preconditioners in its place.

\begin{algorithm}[t]
    \caption{
        $L^p$ Adam~\citep{kingma2015adam}.
        All vector operations are element wise.
        We found that $p > 2$ works well in practice.
        Recommended default hyperparameters:
        $\beta_1=0.9$, $\beta_2=0.999$, $\epsilon=10^{-8}$.
        The choice of $\alpha$ may depend on $p$;
        we recommend starting with
        ${ \alpha = 10^{-(1.5 + \slashdiv{3}{p})} }$
        to achieve Adam's usual step size bound of $10^{-2.5}$.
    }
    \label{algo:adam_lp}
    \setstretch{1.25}
    \begin{algorithmic}
        \State Select $L^p$ norm ($p > 0$)
        \State Select hyperparameters $\alpha > 0$, $\beta_1 \in [0,1)$, $\beta_2 \in (\beta_1^p,1)$, $\epsilon > 0$
        \State Initialize parameters $\theta_0 \in \mathbb{R}^d$ for stochastic objective function $f(\theta)$
        \State Initialize $m_0 \gets 0$, $v_0 \gets 0$
        \For{$t = 1\ \text{to}\ T$}
            \State $g_t \gets \nabla_\theta f_t(\theta_{t-1})$
            \State $m_t \gets \beta_1 \cdot m_{t-1} + (1-\beta_1) \cdot g_t$
            \State $\hat{m}_t \gets m_t \mathbin{/} (1 - \beta_1^t)$
            \State $v_t \gets \beta_2 \cdot v_{t-1} + (1-\beta_2) \cdot \abs{g_t}^p$
            \State $\hat{v}_t \gets v_t \mathbin{/} (1 - \beta_2^t)$
            \State $\theta_t \gets \theta_{t-1} - \alpha \cdot \hat{m}_t \mathbin{/} (\epsilon + \nroot{p}{\hat{v}_t})$
        \EndFor
        \State \Return $\theta_T$
    \end{algorithmic}
\end{algorithm}

\tightsection{Analysis}
\label{sect:analysis}

RMSProp and Adam normalize their optimization steps by the exponentially weighted RMS of gradients, which can be interpreted as a moving approximation of AdaGrad's preconditioner.
Yet, this particular choice of diagonal rescaling represents just one possibility from an infinite spectrum of related methods based on the $L^p$ norm~\citep{kingma2015adam}.
In this section, we aim to elucidate the mathematical properties of these different norms.
While we focus our discussion on Adam, our results also include RMSProp as a specific case where $\beta_1 = 0$.

The RMS can be viewed as a particular instance of the generalized mean
$\smash{ \nroot{p}{ \frac{1}{n}\sum_{i=1}^n \abs{x_i}^p} }$ of values $x_1, \ldots, x_n$ for $p=2$.
We can therefore rewrite the Adam step in~(\ref{eq:adam}) as
\begin{equation}
    \label{eq:adam_lp}
    - \alpha \cdot \frac{m_t}{\nroot{p}{v_t}}
    \quad \text{($L^p$ Adam)}
\end{equation}
where $v_t \gets \beta_2 \cdot v_{t-1} + (1-\beta_2) \cdot \abs{g_t}^p$.
For clarity, we have omitted initialization bias correction~\citep{kingma2015adam} and the small constant $\epsilon > 0$ added to the denominator for numerical stability.
The pseudocode in Algorithm~\ref{algo:adam_lp} includes these additional details for reference.

The $L^p$ Adam update was presented by~\cite{kingma2015adam} as a theoretical vehicle for obtaining the AdaMax algorithm, by taking the limit as $p \to \infty$.\footnote{
    To induce proper convergence of the limit to generate AdaMax, $\beta_2$ must be reparameterized as $\beta_2^p$.
}
No other theoretical or empirical analysis has been conducted to our knowledge,
and it is therefore unknown what implications are held by choices of finite $p$ other than $2$.
We illuminate these implications in the remainder of this section by studying how Adam's properties are affected (or unaffected) when the $L^p$ norm is changed.

\tightparagraph{Positive definiteness}
Recall from Section~\ref{sect:background} that element-wise division can be interpreted as left multiplication by the inverse of a diagonal matrix.
This inverse can be understood as a preconditioner $P \in \mathbb{R}^{d \times d}$ that (ideally) makes the objective function more amenable to minimization with SGD:
$\theta_t \gets \theta_{t-1} - \alpha \cdot P g_t$.
It is this view that underpins the theoretical motivation behind AdaGrad, and from which Adam derives its use of RMS normalization.
In general, any positive definite matrix can be used for $P$, since this guarantees that the expected step direction $\mathbb{E}[P g_t]$ is a descent direction.
For $L^p$ Adam, we have that $\smash{ P = \diag(\slashdiv{1}{\nroot{p}{v_t}}) }$, which is guaranteed to be positive definite since it is diagonal with positive values.\footnote{
    We can safely ignore the case where $\nroot{p}{v_t} = 0$ because of the constant $\epsilon>0$ that is added to Adam's denominator in practice. See Algorithm~\ref{algo:adam_lp}.
}
While this does not mean that every $L^p$ norm yields a \textit{good} preconditioner, the perspective of $L^p$ Adam as a form of preconditioned SGD assures us that substituting non-Euclidean norms (i.e.\ $p \neq 2$) is justified.

\tightparagraph{Scale invariance}
A useful property of Adam is that its steps are unaffected by a diagonal rescaling of the gradients.
To see this, suppose all gradients are multiplied by a constant vector $c$ whose elements are positive.
Then, from (\ref{eq:adam_lp}), we have
\begin{equation}
    - \alpha \cdot \frac{c \cdot m_t}{\nroot{p}{c^p \cdot v_t}}
    = - \alpha \cdot \frac{m_t}{\nroot{p}{v_t}}
\end{equation}
and hence any choice of $L^p$ norm preserves scale invariance.
This is important in practice, since changes in the network topology can dramatically alter the scale of the partial derivatives.

\tightparagraph{Maximum step size}
Another property of Adam that enables it to train a wide range of neural architectures is its bounded step size, or ``trust region'' effect~\citep{kingma2015adam}.
Adam takes its largest step when it experiences a sudden nonzero gradient $g_t$ after a long sequence of zero gradients.
In this case, we have that $m_{t-1} \to 0$ and $v_{t-1} \to 0$, and so the magnitude of the step becomes
\begin{equation}
    \label{eq:max_step}
    \lim_{\substack{m_{t-1} \to 0\\v_{t-1} \to 0}}
    \abs{- \alpha \frac{m_t}{\nroot{p}{v_t}}}
    = \alpha \frac{(1-\beta_1) \cdot \abs{g_t}}{\nroot{p}{(1-\beta_2) \cdot \abs{g_t}^p}}
    = \alpha \frac{1-\beta_1}{\nroot{p}{1-\beta_2}}
\end{equation}
For $p=2$ with typical hyperparameter values $\beta_1 = 0.9$ and $\beta_2 = 0.999$, this quantity is $\sqrt{10} \alpha$, meaning step sizes are roughly bounded by triple the learning rate in all dimensions.
We can see from~(\ref{eq:max_step}) that the step size bound is highly dependent on the choice of $L^p$ norm.
For example, if one naively chooses $p=1$ with these same hyperparameters, then the maximum step magnitude would become a massive $100 \alpha$.
It is also interesting to note that there is a lower bound on the maximum step size;
letting $p$ become arbitrarily large in~(\ref{eq:max_step}), we arrive at the conclusion that the maximum step can never be less than $\alpha (1-\beta_1)$, or $0.1 \alpha$ for the default $\beta_1=0.9$.
\begin{figure}[t]
    \centerline{
        \includegraphics[width=0.45\textwidth]{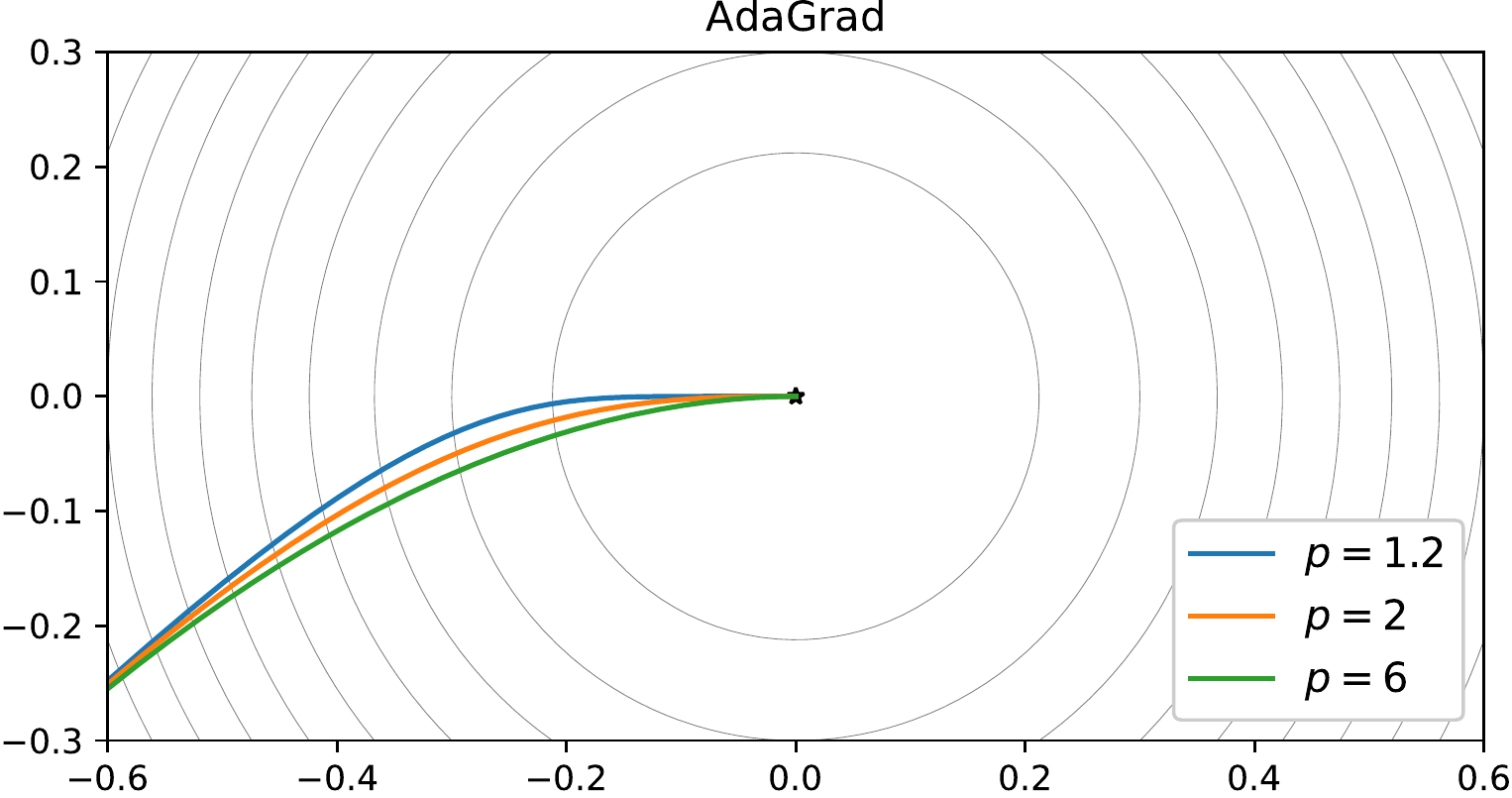}
        \hfill
        \includegraphics[width=0.45\textwidth]{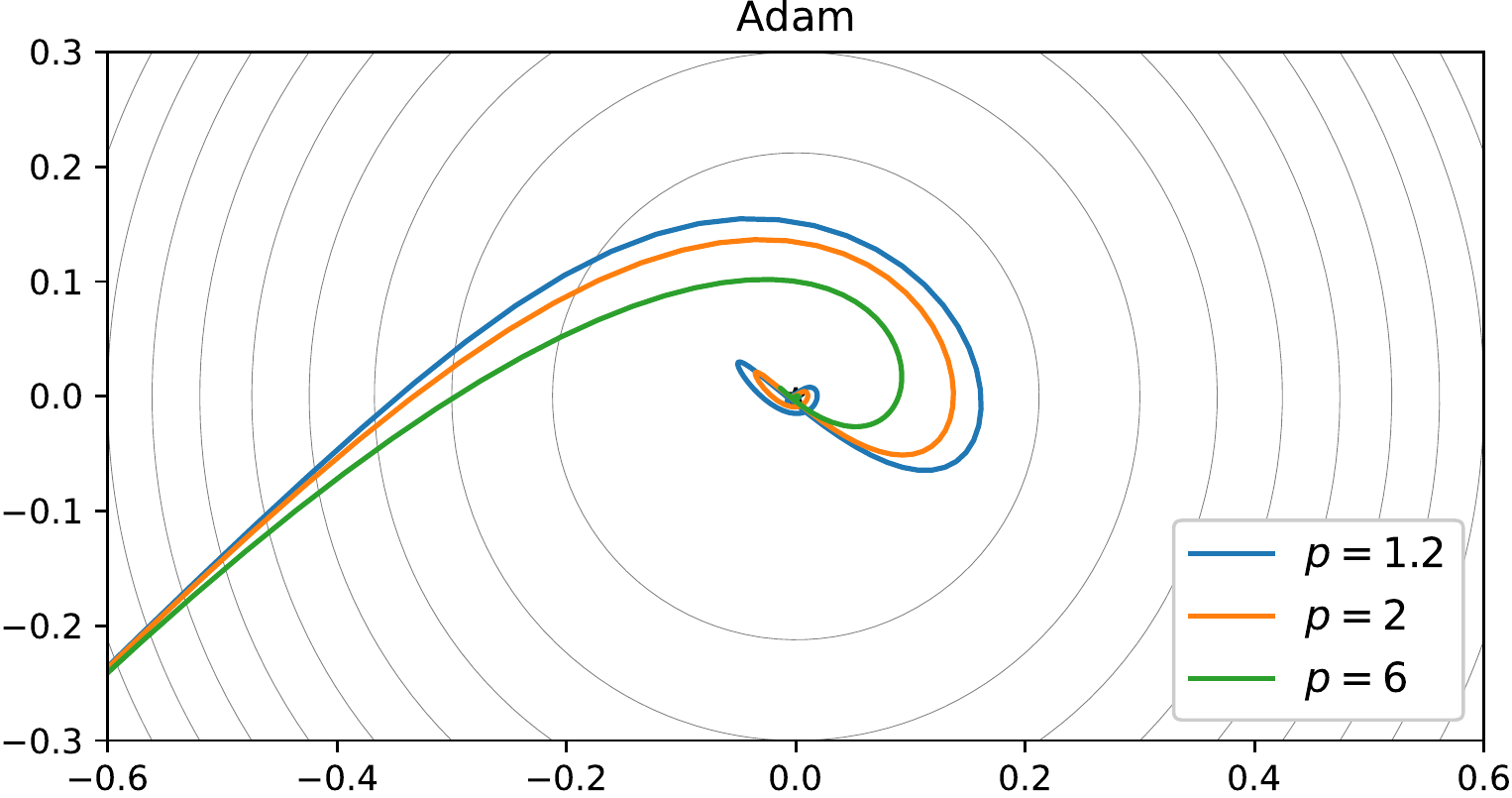}
    }
    \figspace
    \centerline{
        \includegraphics[width=0.45\textwidth]{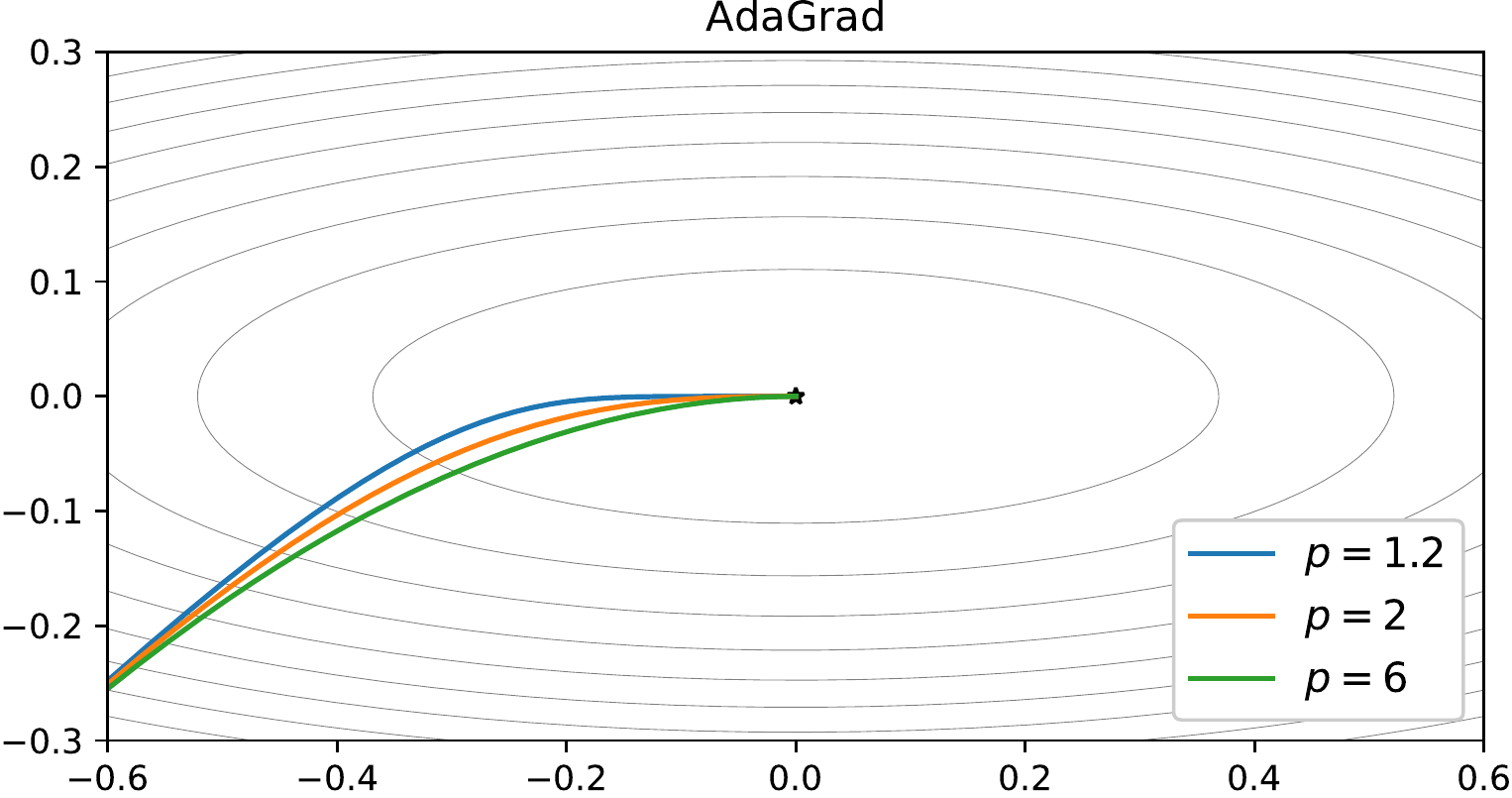}
        \hfill
        \includegraphics[width=0.45\textwidth]{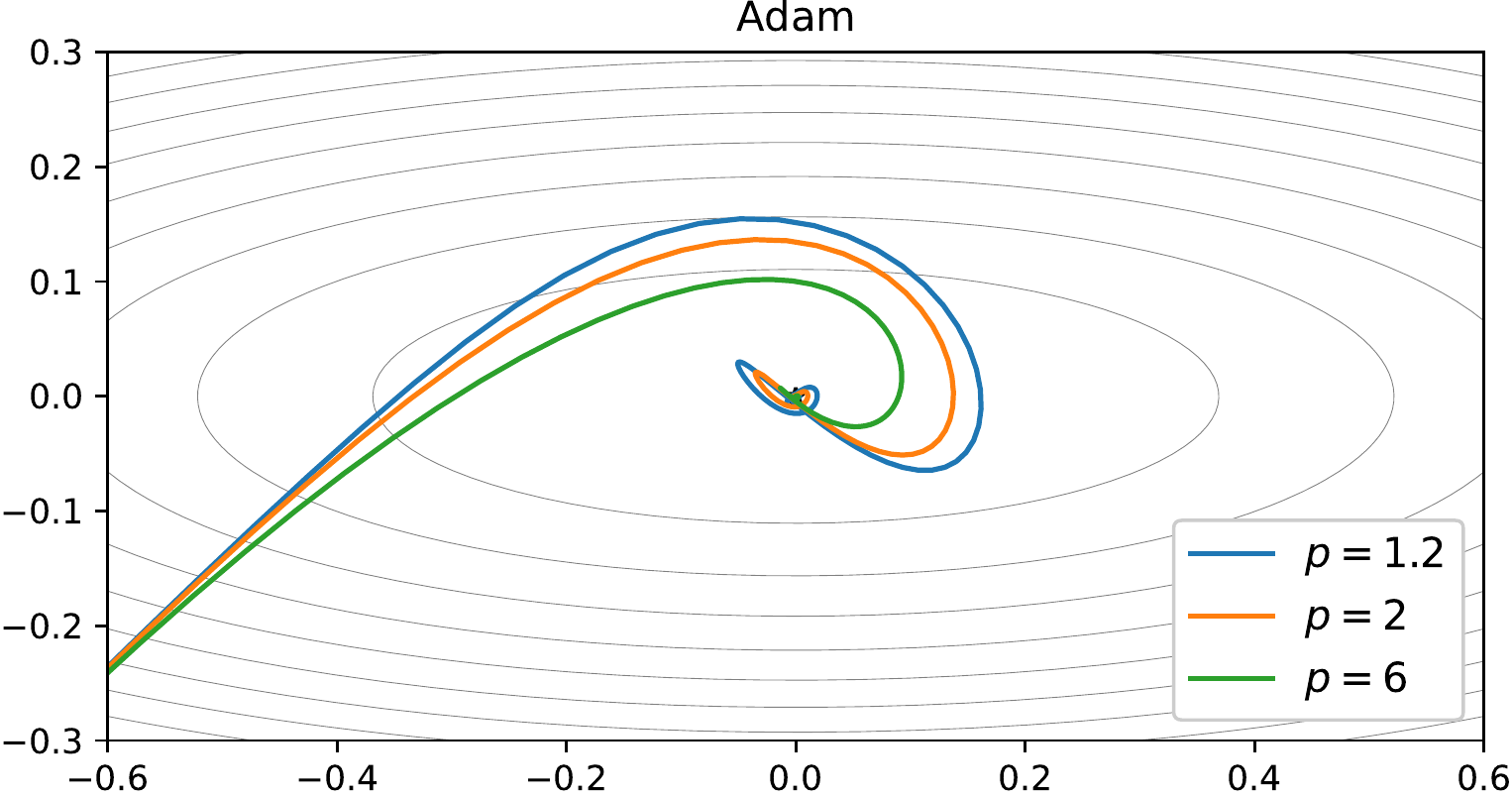}
    }
    \caption{
        Visualizations of the paths taken by $L^p$-variants of AdaGrad and Adam for $p \in \{1.2, 2, 6\}$ and $\alpha = 0.05$ when minimizing two quadratic functions, starting from the point
        $\smash{ ( - \frac{\sqrt{3}}{2}, - \frac{1}{2} ) }$.\\
        Top row:
        $f(x,y) = \frac{1}{2} \left( x^2 + y^2 \right)$.
        Bottom row:
        $f(x,y) = \frac{1}{2} \left( \frac{9}{100} x^2 + y^2 \right)$.
        The gradients of these functions differ only by a diagonal rescaling, so the methods' paths are unchanged.
    }
    \label{fig:quadratics}
\end{figure}

\tightparagraph{Typical step size}
The maximum step analysis is useful for providing guarantees about training stability, but it represents a worst-case scenario.
We would also like to understand how the typical step of Adam is affected by $p$.
While computing this quantity directly is not possible without precise knowledge of the distribution of $g_t$, we can still infer a trend.
In general, if we have $q > p$, then the $L^q$ norm will be less than or equal to the $L^p$ norm~\citep{rolewicz1972metric};
however, we must consider what happens under Adam's exponential weights.
Letting $w_i = (1-\beta_2) \beta_2^{t-i}$, Adam's denominator at iteration $t$ can be rewritten as
\begin{align}
    \allowdisplaybreaks
    \nroot{p}{\sum\nolimits_{i=1}^t w_i \cdot \abs{g_t}^p}
    &= \nroot{p}{\sum\nolimits_{i=1}^t \abs{w_i^\slashdiv{1}{p} \cdot g_t}^p} \\
    \label{eq:ineq1}
    &\geq \nroot{q}{\sum\nolimits_{i=1}^t \abs{w_i^\slashdiv{1}{p} \cdot g_t}^q}
    = \nroot{q}{\sum\nolimits_{i=1}^t w_i^\slashdiv{q}{p} \cdot \abs{g_t}^q}
\end{align}
We see that the weighted $L^p$ norm is at least as large as a re-weighted $L^q$ norm.
Since $w_i < 1$, then $\smash{ w_i^\slashdiv{q}{p} < w_i }$, so it is not possible to lower-bound this quantity by the weighted $L^q$ norm.
Hence, the norm does not decrease monotonically under exponential weights, which means Adam's steps do not increase monotonically with $p$;
this makes sense, since our preceding analysis showed that the maximum step \textit{decreases} with larger $p$, which would be contradictory.

\paragraph{Inertia}
Unlike RMSProp, Adam utilizes momentum that helps it move through regions of the parameter space where the gradient is close to zero (``plateaus'').
We call this \textit{inertia}, referring to the momentum's tendency to continue proceeding, or ``drifting,'' in the absence of a gradient signal.
One might wonder whether this ability is adversely affected by a different choice of $L^p$-norm.
The analysis for this case is essentially the opposite of the maximum step analysis;
suppose that after the first $t$ iterations of training, all encountered gradients are suddenly zero.
By induction, we can determine that $m_{t+i} = \beta_1^i \cdot m_t$ and $v_{t+i} = \beta_2^i \cdot v_t$.
Substituting these into (\ref{eq:adam_lp}), we have that the step at iteration $t+i$ will be
\begin{equation}
    \label{eq:inertia}
    - \alpha \cdot \frac{\beta_1^i \cdot m_t}{\nroot{p}{\beta_2^i \cdot v_t}}
    = - \alpha \cdot \frac{m_t}{\nroot{p}{v_t}} \cdot \left( \frac{\beta_1}{\nroot{p}{\beta_2}} \right)^i
    \approx - \alpha \cdot \frac{m_t}{\nroot{p}{v_t}} \cdot \beta_1^i
\end{equation}
Hence, the optimization proceeds in the same direction, but with a magnitude that decays\footnote{
    Intriguingly, equation~(\ref{eq:inertia}) implies a constraint on $L^p$ Adam's hyperparameters:
    we must have $\beta_1 < \nroot{p}{\beta_2}$, or else the step magnitude will not decay.
    We have included this constraint in Algorithm~\ref{algo:adam_lp}.
}
exponentially at a rate of $\beta_1 \mathbin{/} \nroot{p}{\beta_2}$.
This value is technically maximized when $p=1$ (assuming $p \geq 1$), but since $\beta_2 \approx 1$ in practice, we obtain the approximate relation in~(\ref{eq:inertia}).
As a result, we can deduce that Adam's inertia is insensitive to different values of $p$.
The significance of this is that changing Adam's norm should not hamper its ability to handle sparse-gradient problems.

\tightparagraph{Summary}
Our analysis has justified the usage of any $L^p$ normalization for Adam by relating it to a positive definite, diagonal preconditioner.
The preconditioner preserves scale invariance and inertia, both of which are important properties for stochastic optimization.
Changing $p$ appears to have an effect primarily on the magnitude bound of the steps taken by Adam;
larger values of $p$ change the size of every step, although not necessarily equally along each dimension, therefore changing the direction as well.
For this reason, its effect is not linear, in contrast to that of the learning rate---although it may still be useful to conceptualize $p$ as an ``inverted'' learning rate.

\begin{table}[b]
    \setlength{\tabcolsep}{12pt}  
    \centering
    \caption{
        The five $L^p$ norms we tested in our experiments, and their corresponding maximum step sizes for Adam when $\beta_1=0.9$ and $\beta_2=0.999$.
        These were chosen to produce maximum step sizes in a geometric sequence with common ratio $\smash{ \sqrt{10} }$, centered around the standard $L^2$ Adam.}
    \begin{tabular}{l *7c}
        \toprule
        \textbf{$L^p$-Norm}     & $1.2$ & $1.5$ & $2$ & $3$ & $6$ \\
        \midrule
        \textbf{Max Step} & $10 \sqrt{10} \alpha$ & $10 \alpha$ & $\sqrt{10} \alpha$ & $\alpha$ & $0.1 \sqrt{10} \alpha$ \\
        \bottomrule
    \end{tabular}
    \label{table:p_values}
\end{table}

\begin{figure}[t]
    \centerline{
        \includegraphics[width=0.333\textwidth]{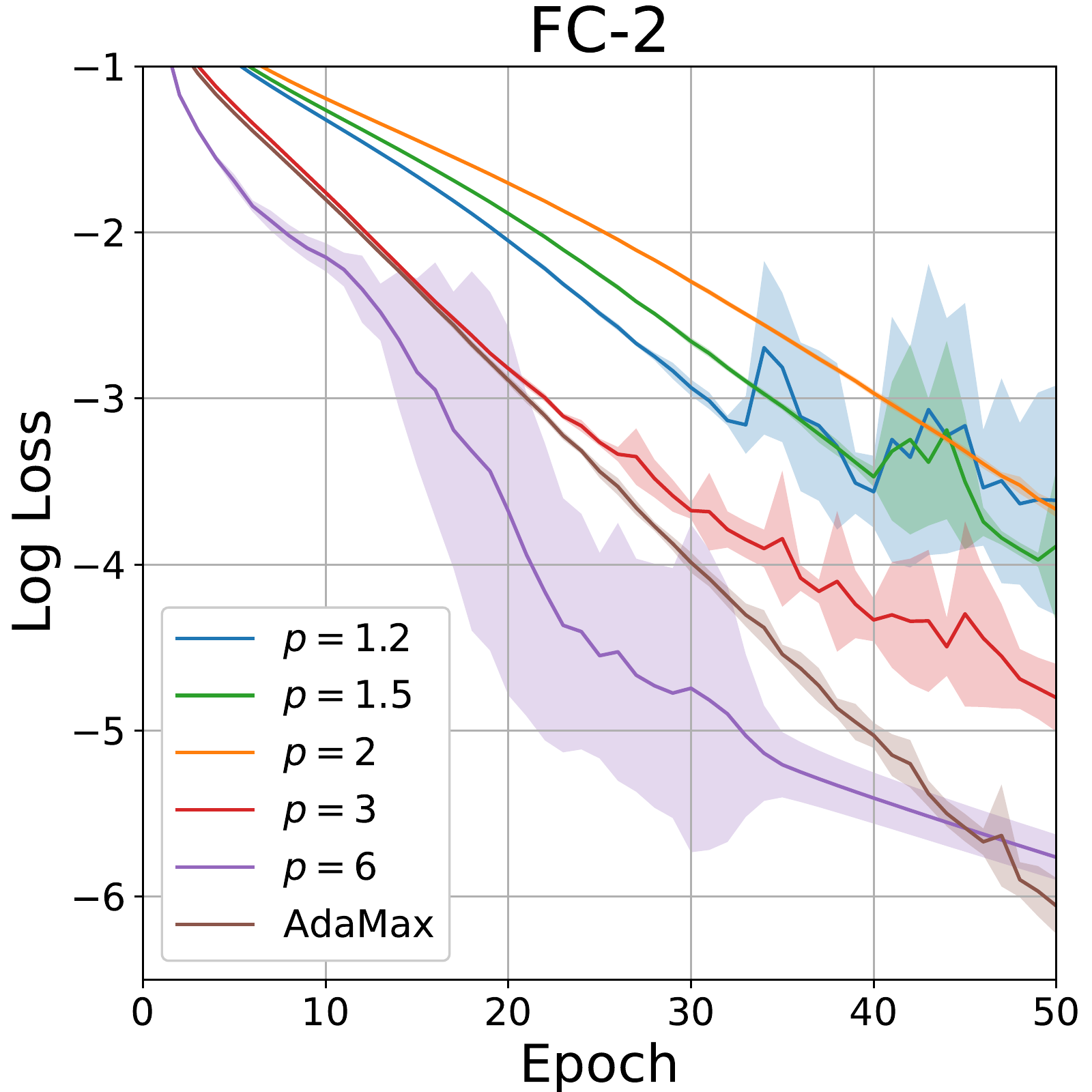}
        \hfill
        \includegraphics[width=0.333\textwidth]{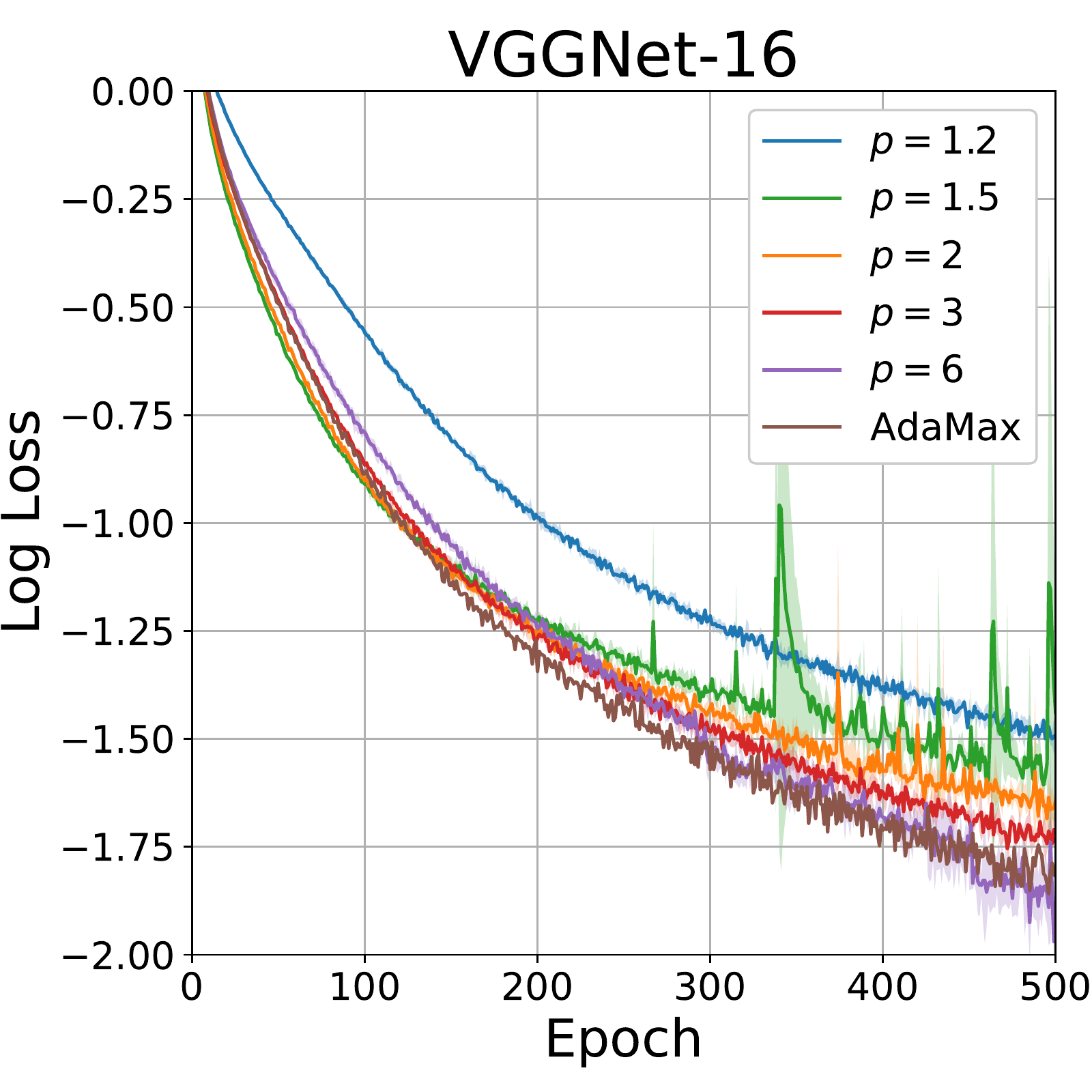}
        \hfill
        \includegraphics[width=0.333\textwidth]{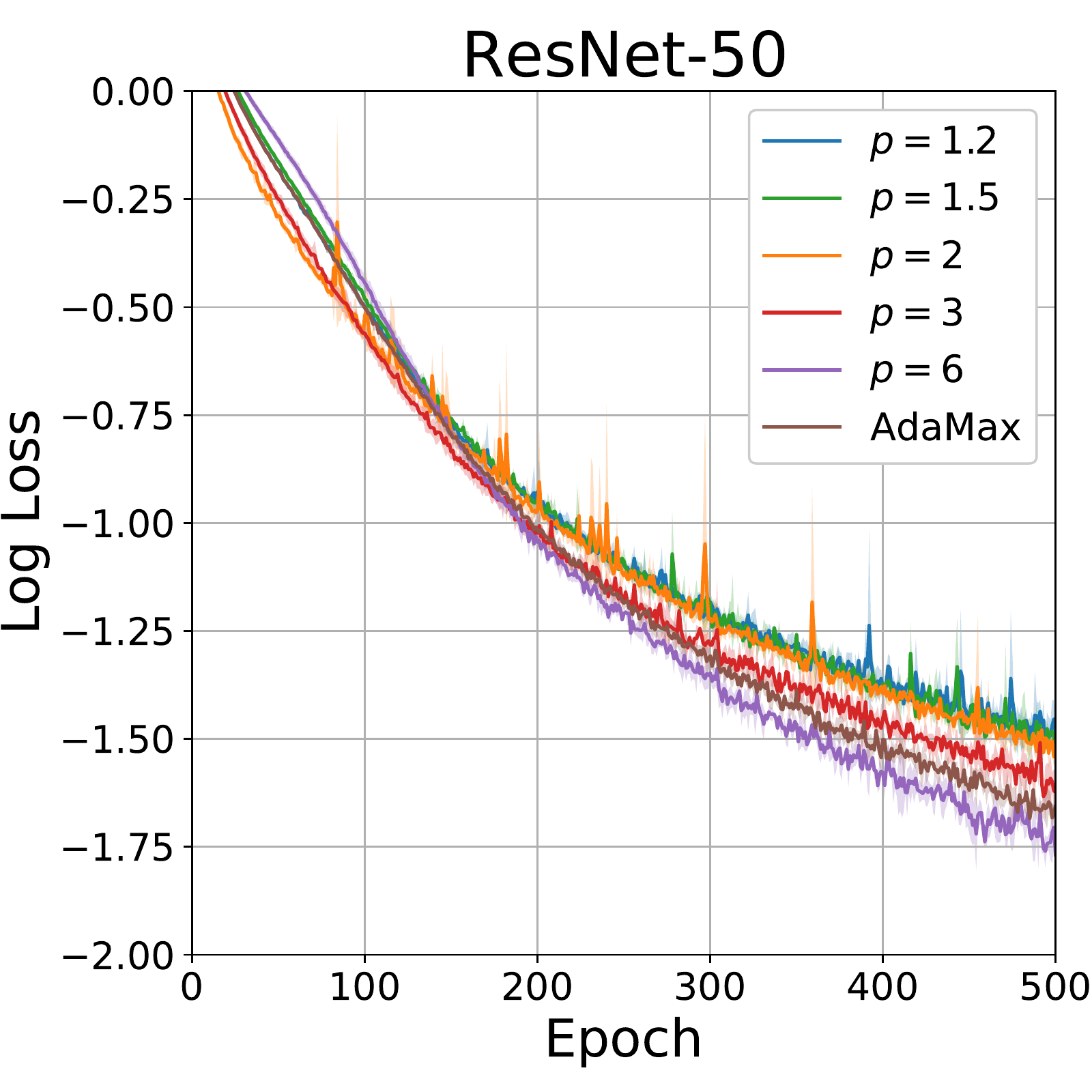}
    }
    \caption{
        Comparison of Adam with various $L^p$ norms on the MNIST and CIFAR-10 classification tasks.
        Results are averaged over 3 trials (10 for MNIST) with standard deviation shaded.
        \label{fig:img_class_lp}
    }
    \figspace
    \centerline{
        \includegraphics[width=0.333\textwidth]{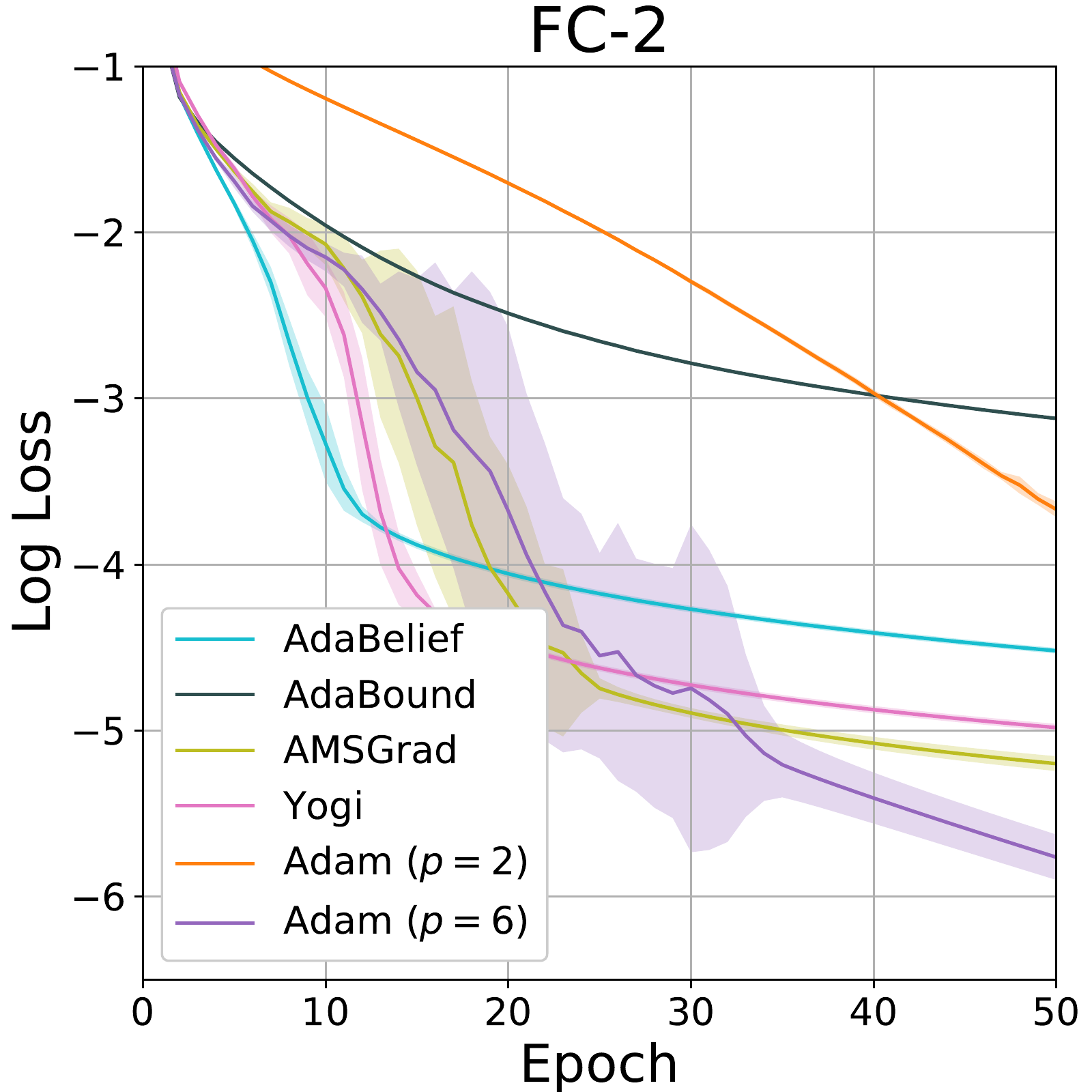}
        \hfill
        \includegraphics[width=0.333\textwidth]{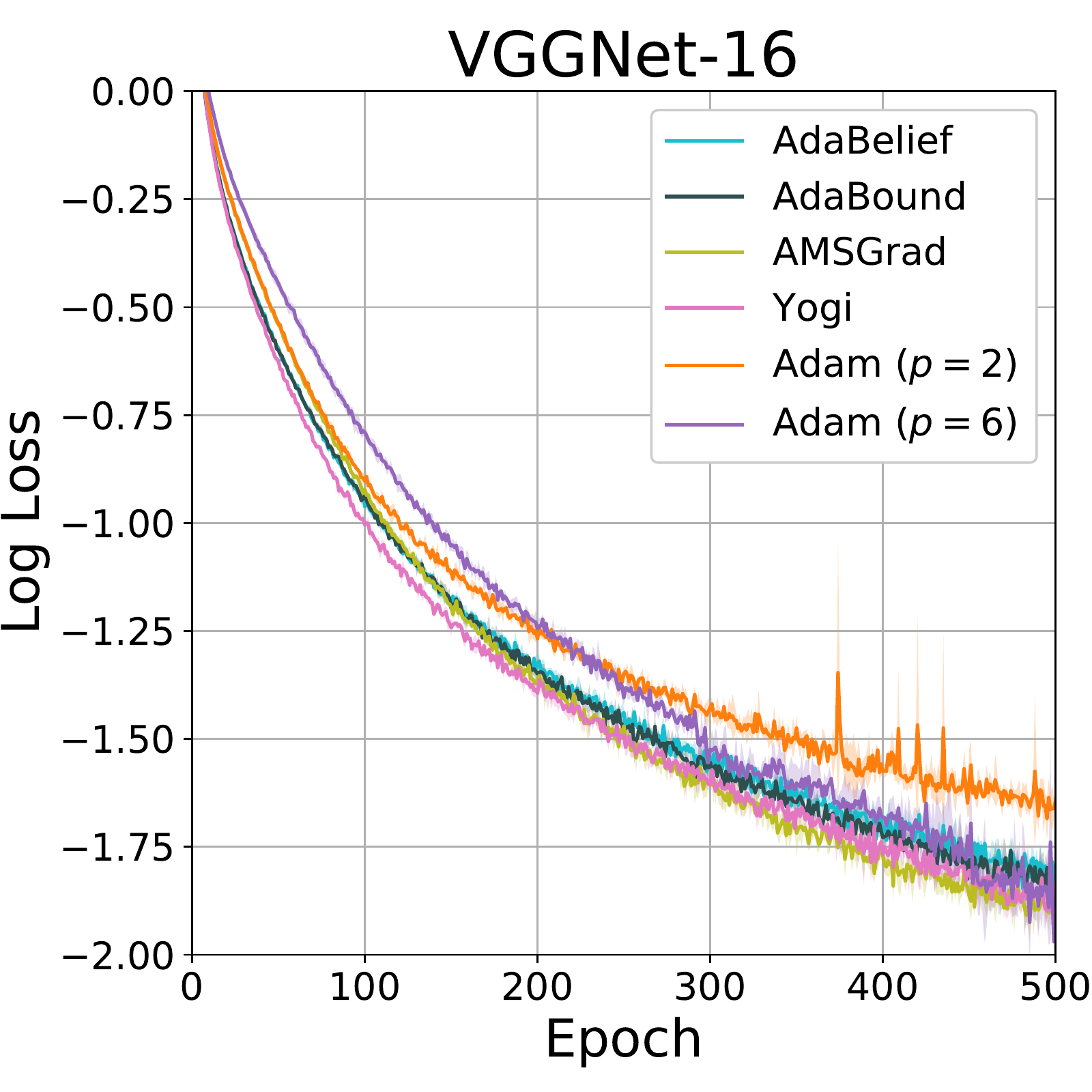}
        \hfill
        \includegraphics[width=0.333\textwidth]{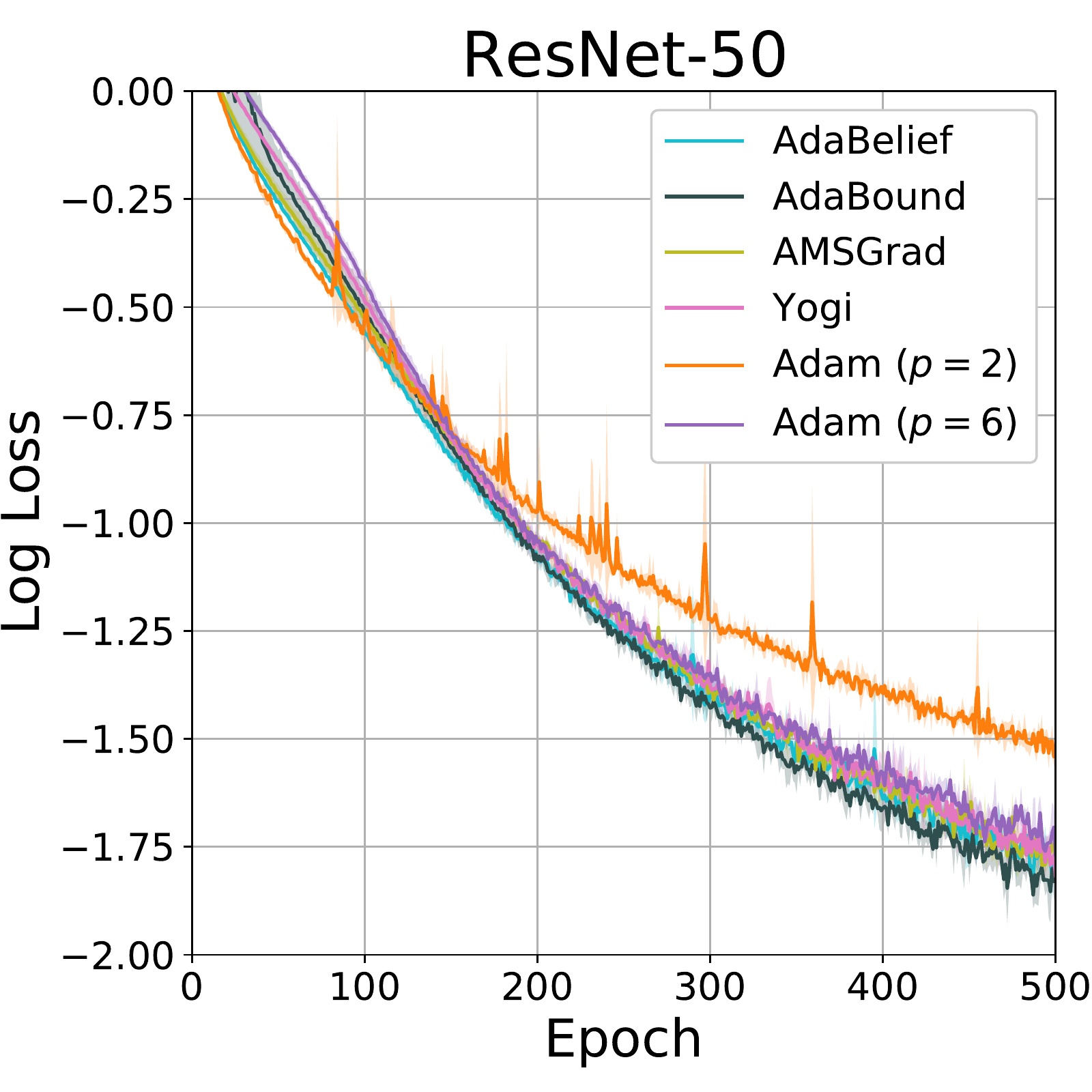}
    }
    \caption{
        Comparison of $L^6$ Adam against four recent methods on the MNIST and CIFAR-10 classification tasks.
        Results are averaged over 3 trials (10 for MNIST) with standard deviation shaded.
        \label{fig:img_class_sota}
    }
\end{figure}

\begin{figure}[t]
    \centerline{
        \includegraphics[width=0.333\textwidth]{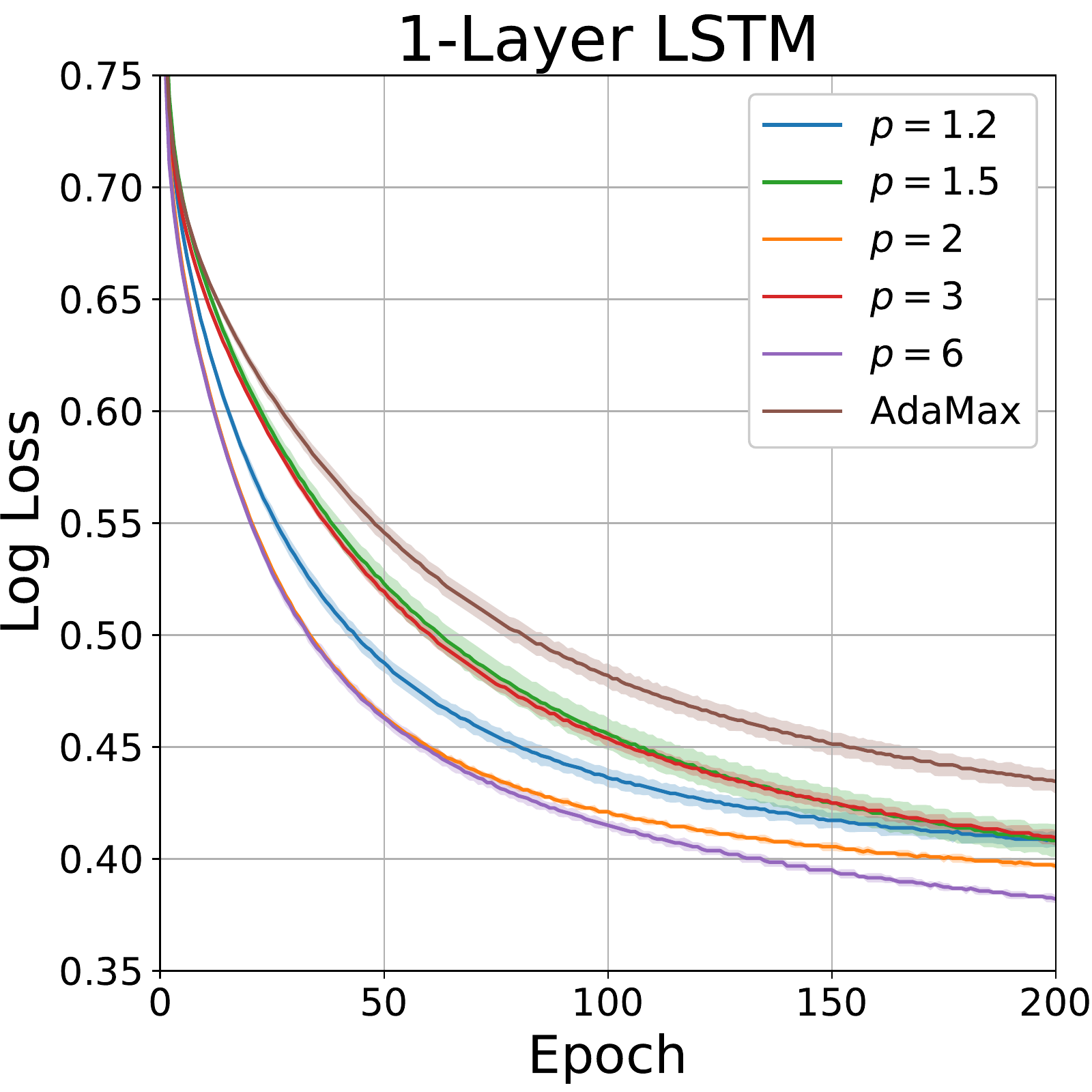}
        \hfill
        \includegraphics[width=0.333\textwidth]{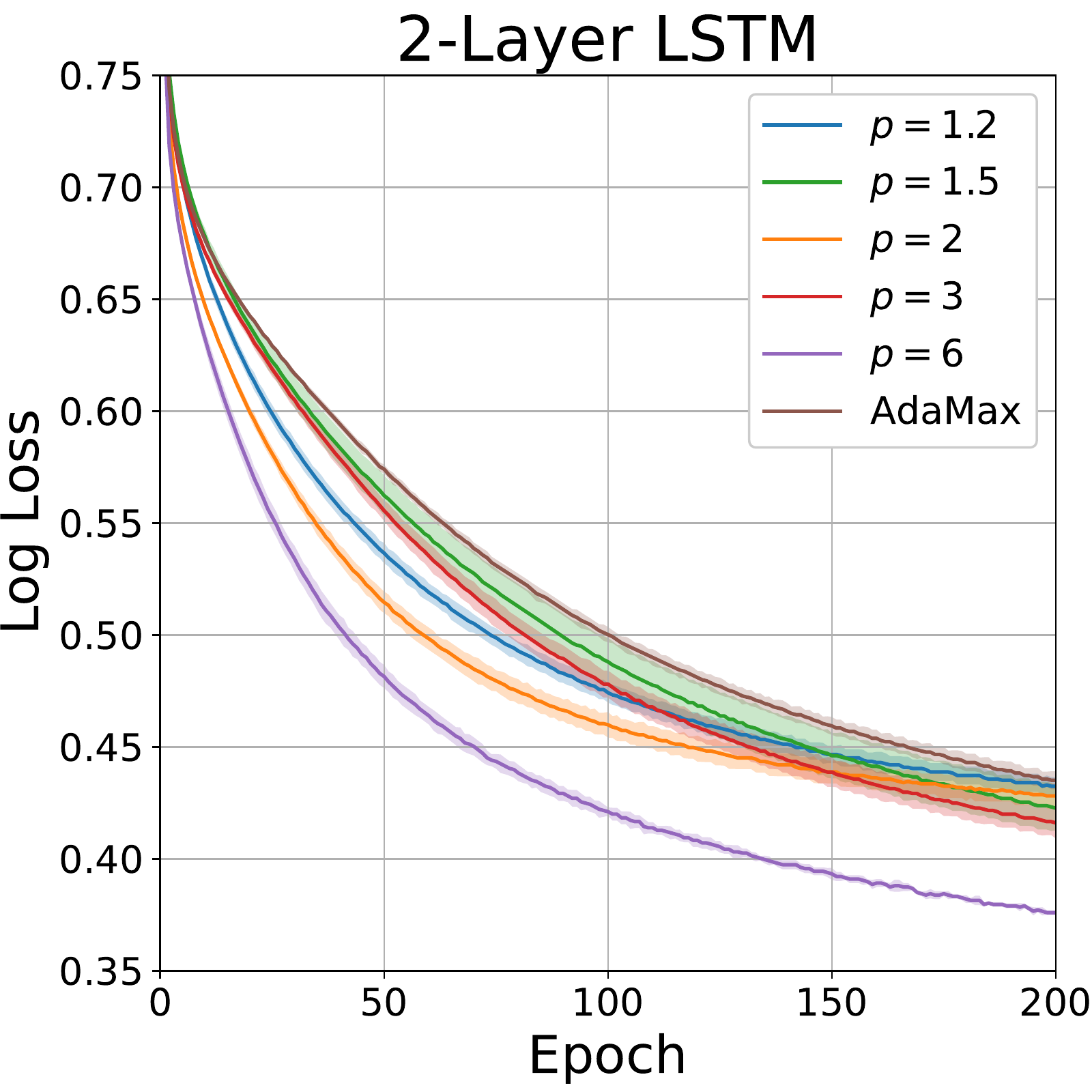}
        \hfill
        \includegraphics[width=0.333\textwidth]{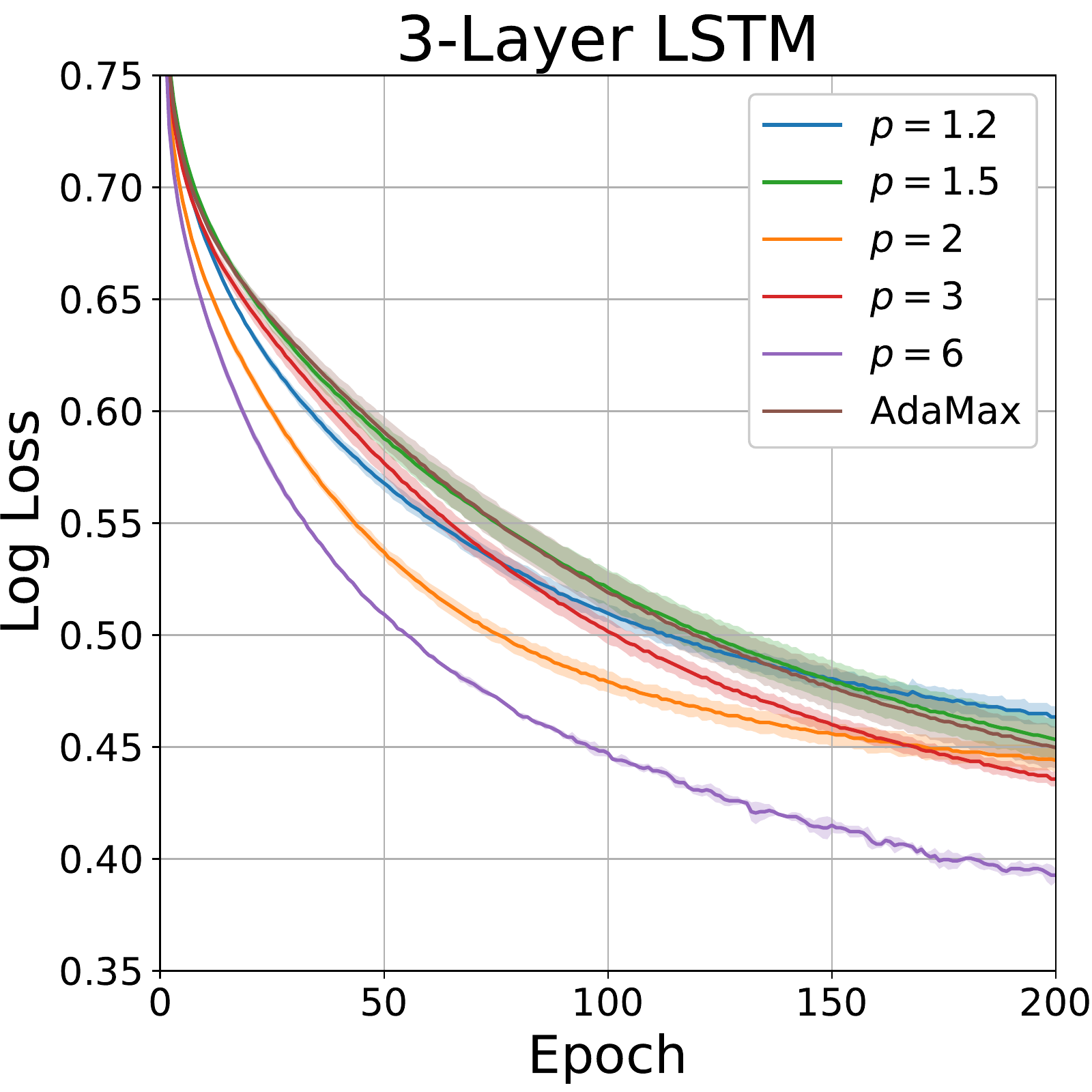}
    }
    \caption{
        Comparison of Adam with various $L^p$ norms on the Penn Treebank language modeling task.
        Results are averaged over 3 trials with standard deviation shaded.
        \label{fig:ptb_lp}
    }
    \figspace
    \centerline{
        \includegraphics[width=0.333\textwidth]{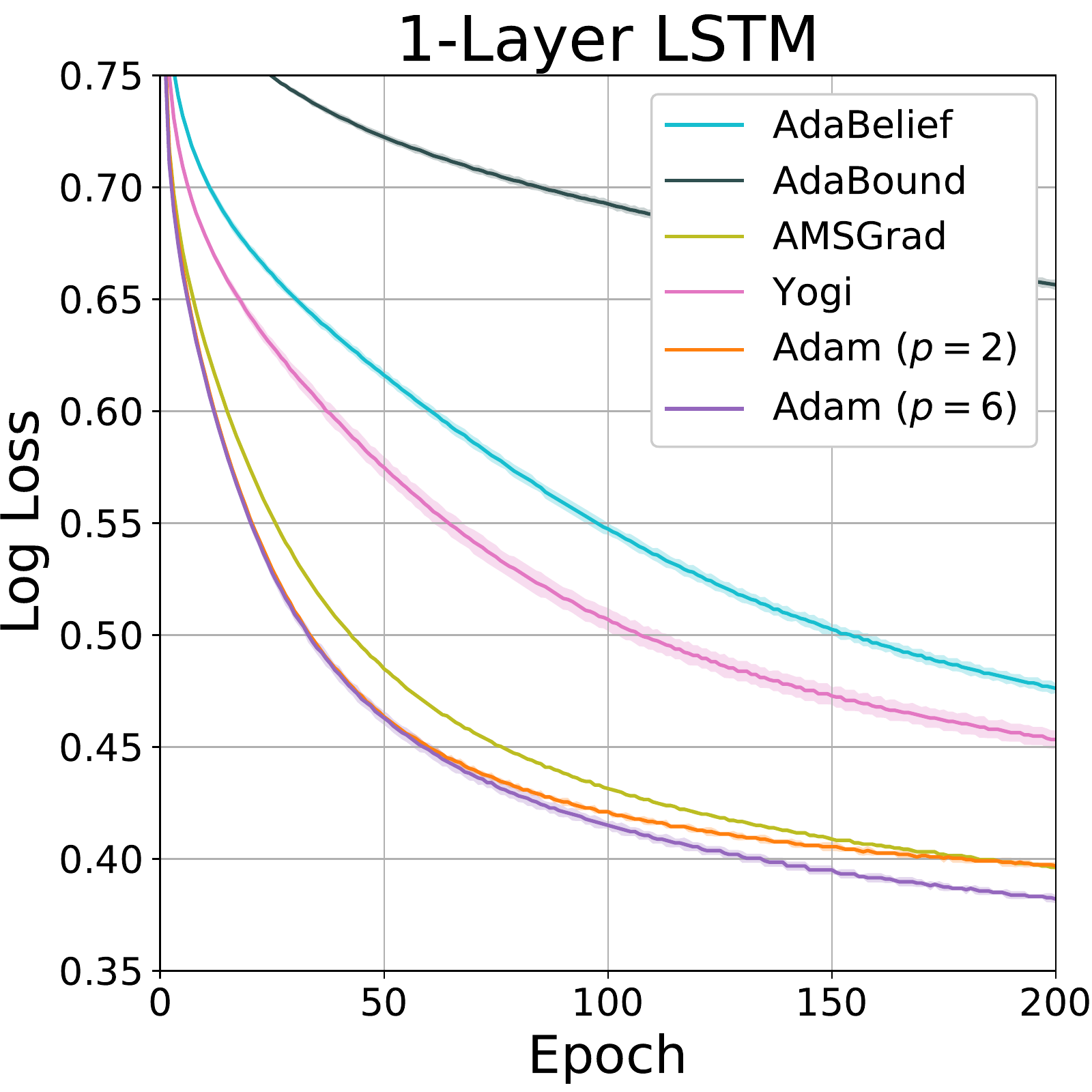}
        \hfill
        \includegraphics[width=0.333\textwidth]{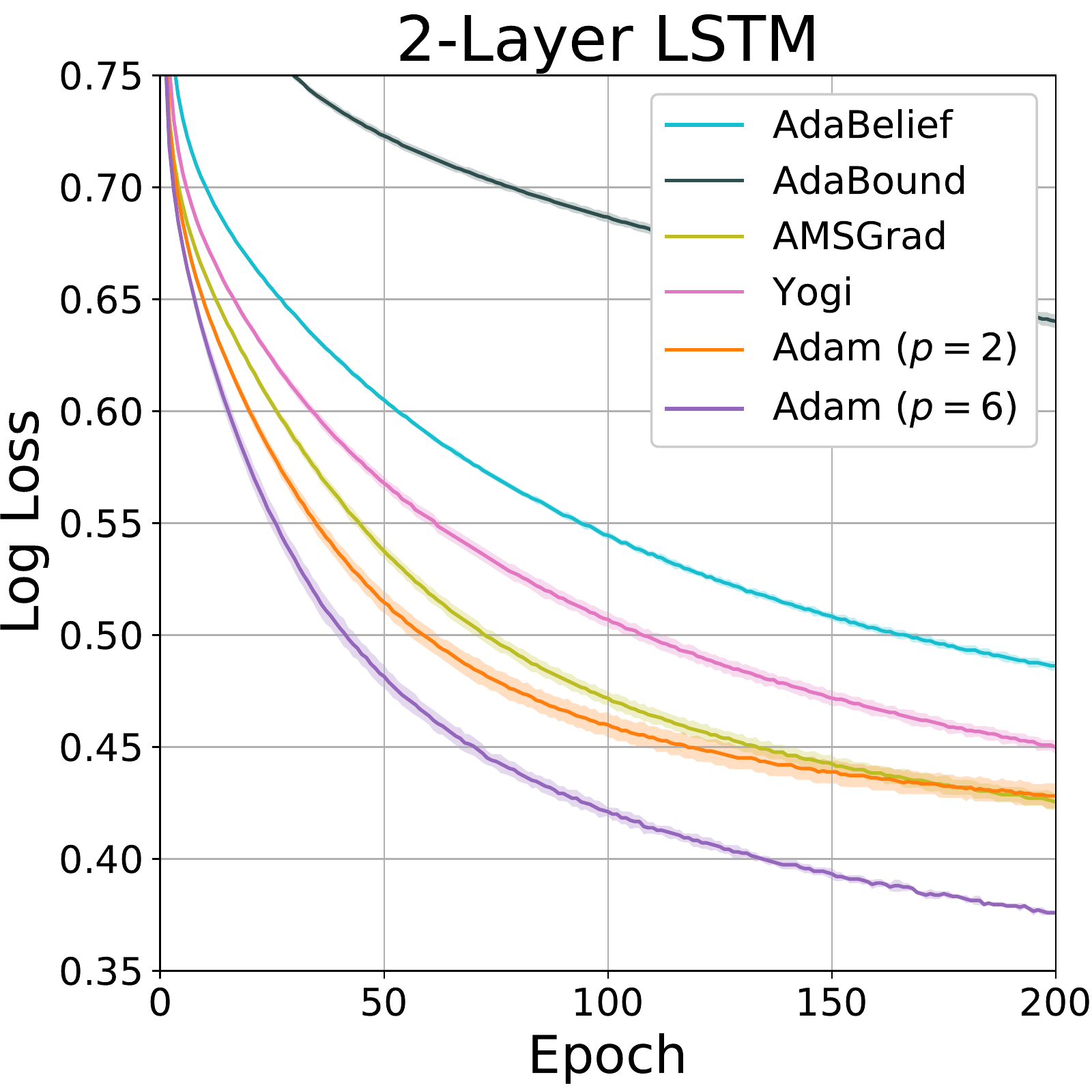}
        \hfill
        \includegraphics[width=0.333\textwidth]{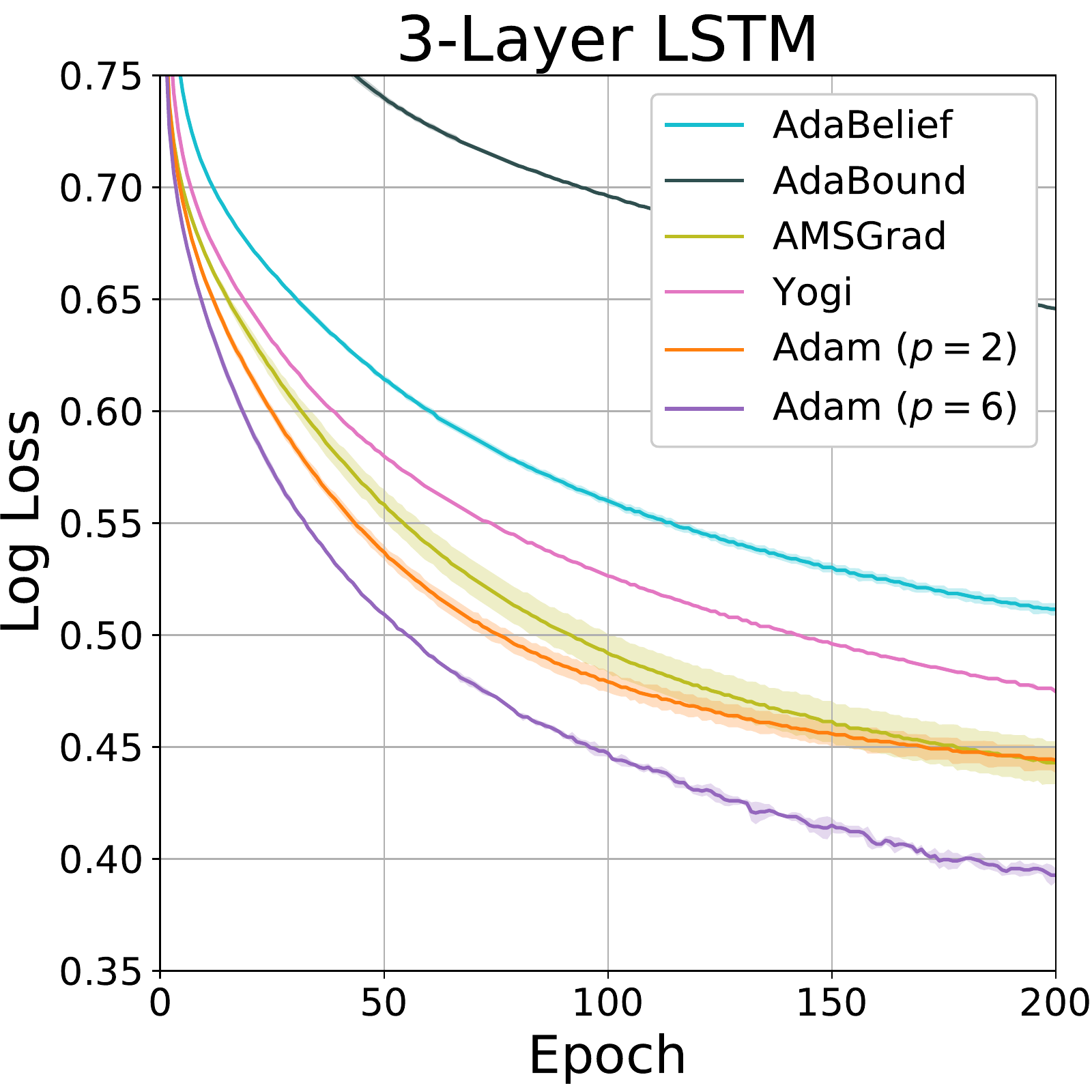}
    }
    \caption{
        Comparison of $L^6$ Adam against four recent methods on the Penn Treebank language modeling task.
        Results are averaged over 3 trials with standard deviation shaded.
        \label{fig:ptb_sota}
    }
\end{figure}

\begin{figure}[t]
    \centerline{
        \includegraphics[width=0.333\textwidth]{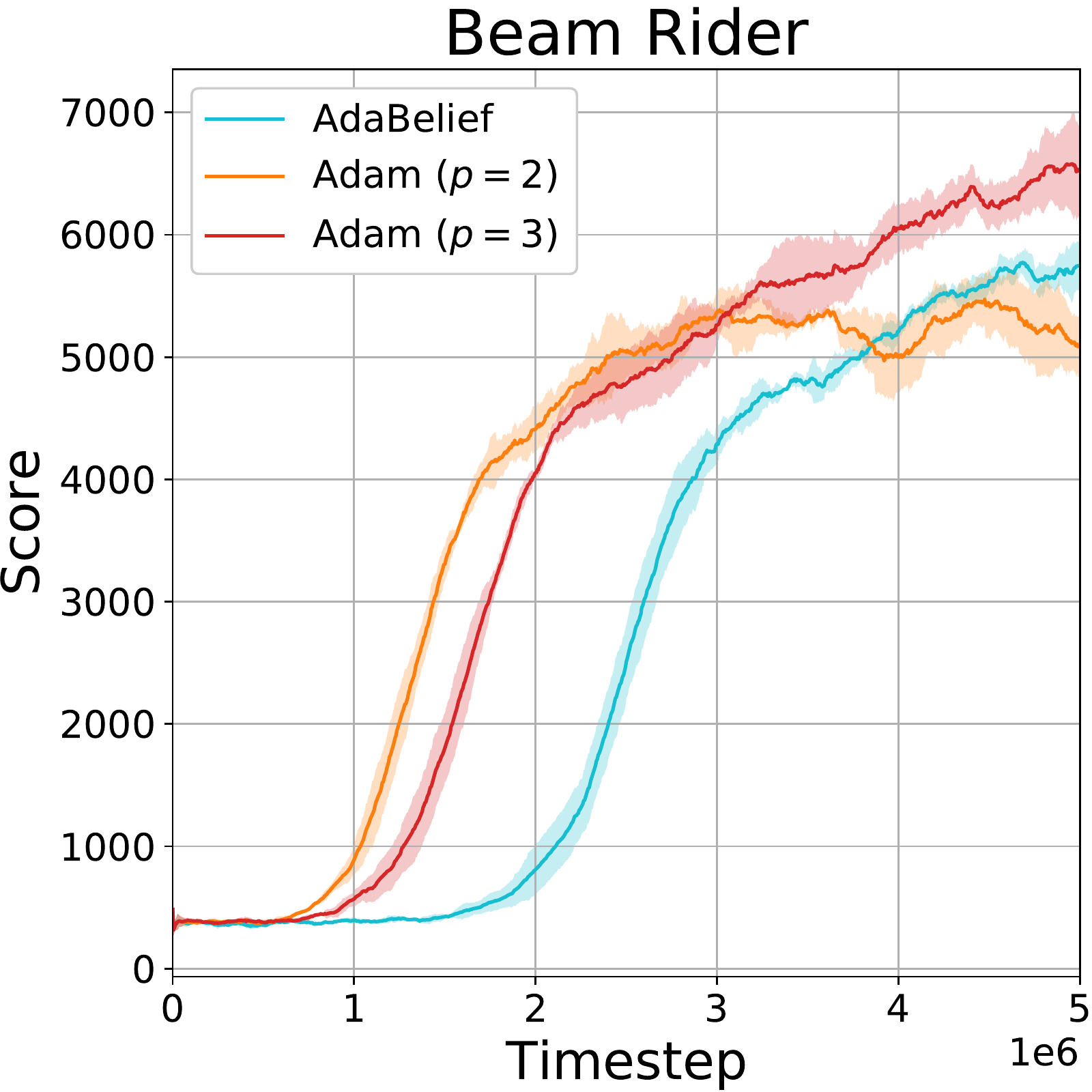}
        \hfill
        \includegraphics[width=0.333\textwidth]{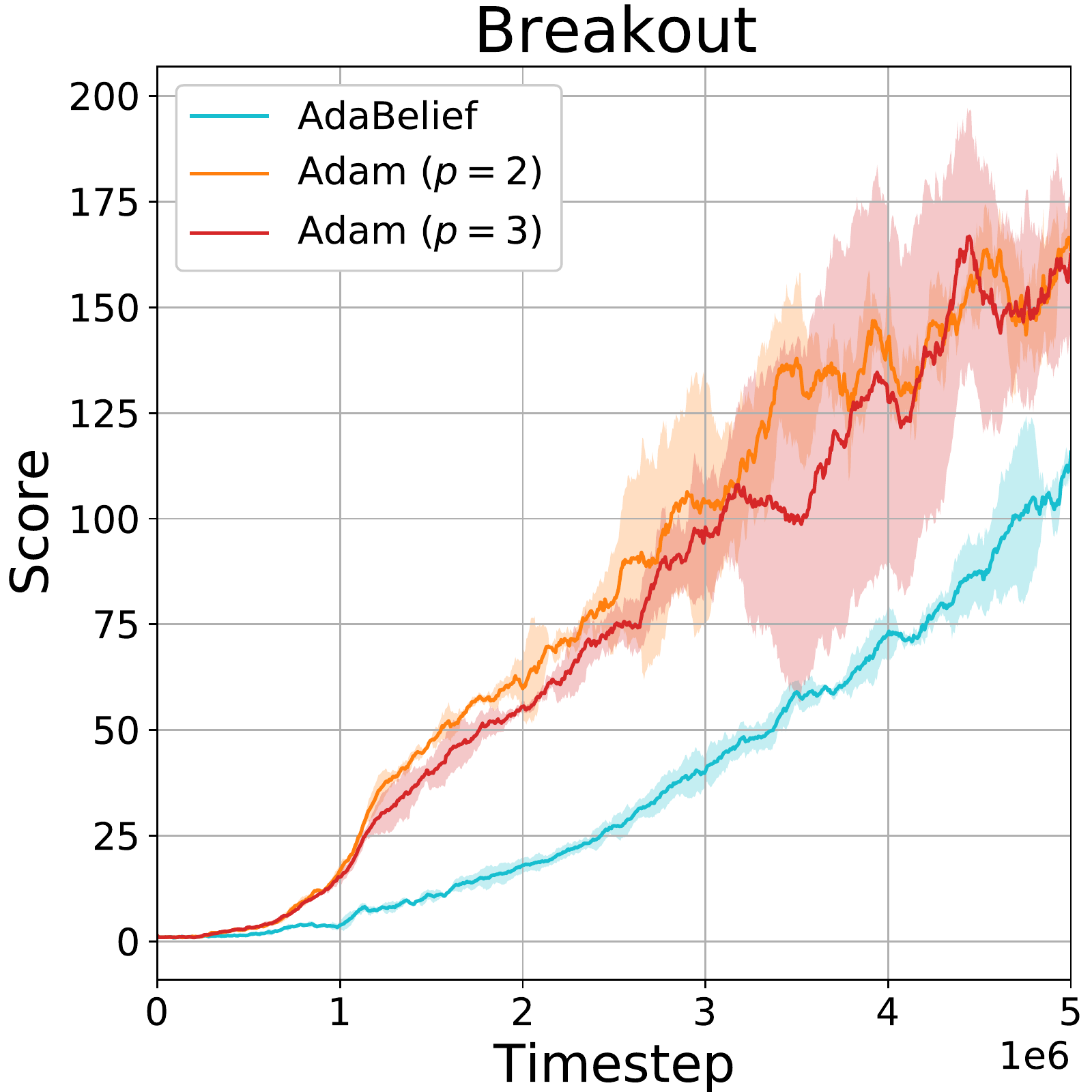}
        \hfill
        \includegraphics[width=0.333\textwidth]{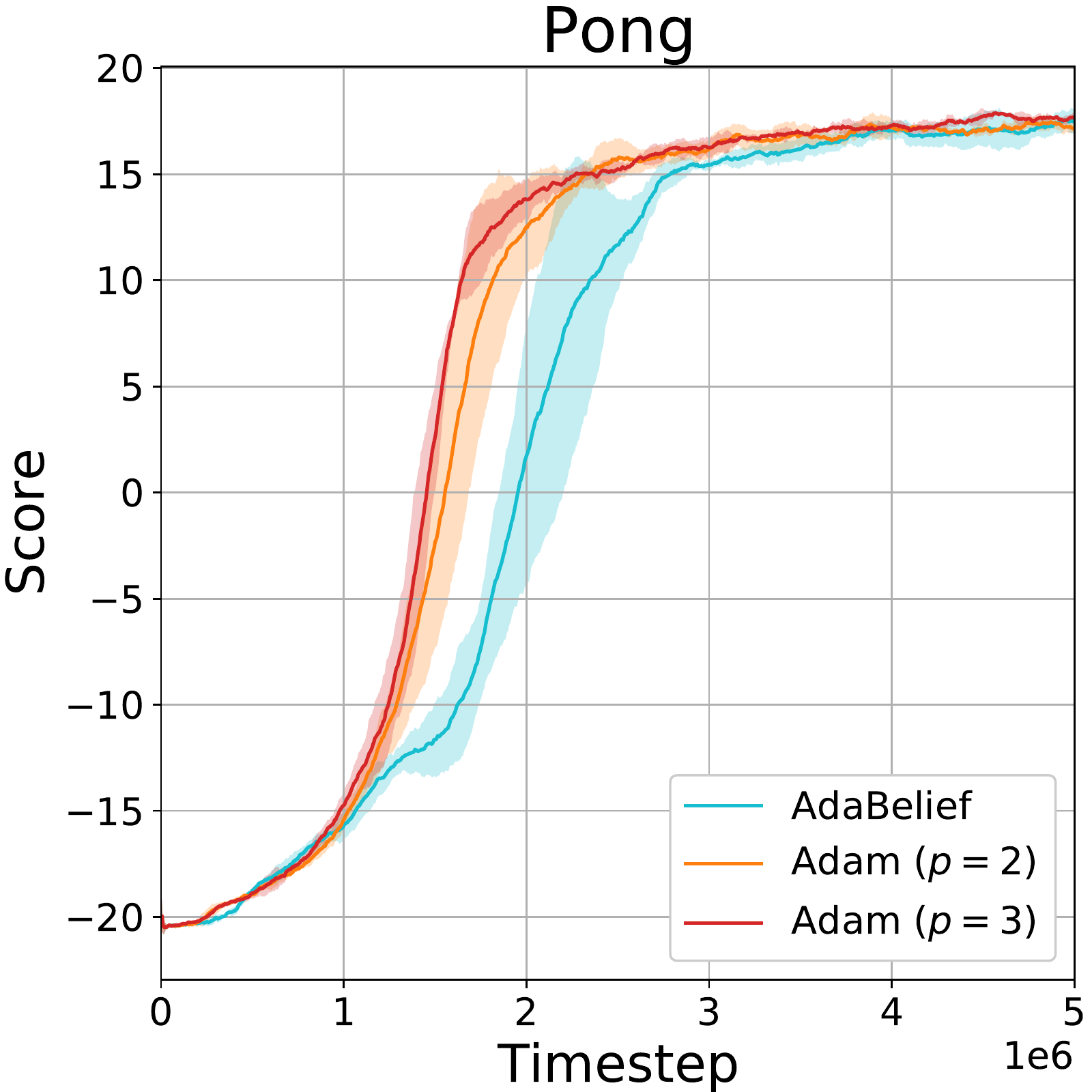}
    }
    \figspace
    \centerline{
        \includegraphics[width=0.333\textwidth]{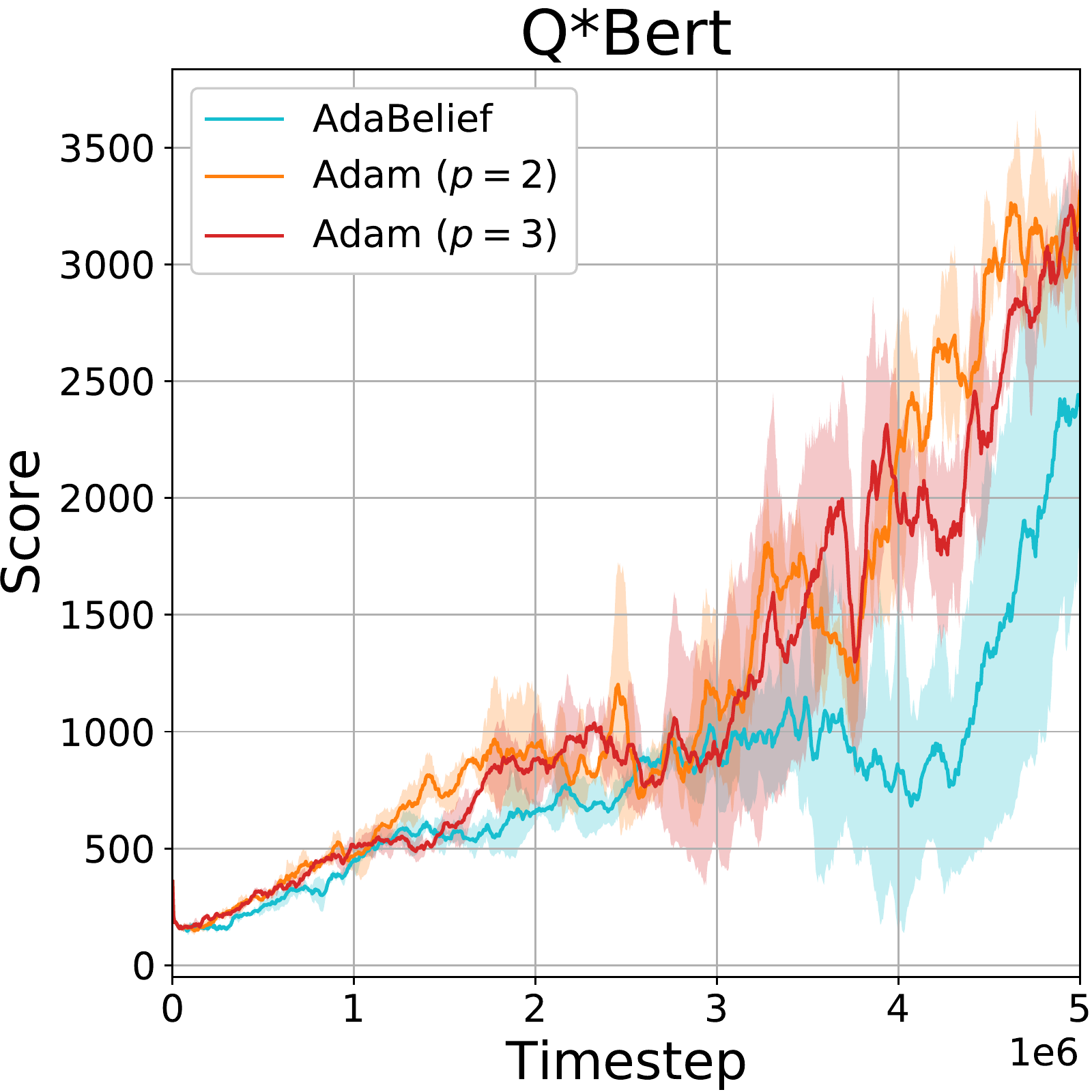}
        \hfill
        \includegraphics[width=0.333\textwidth]{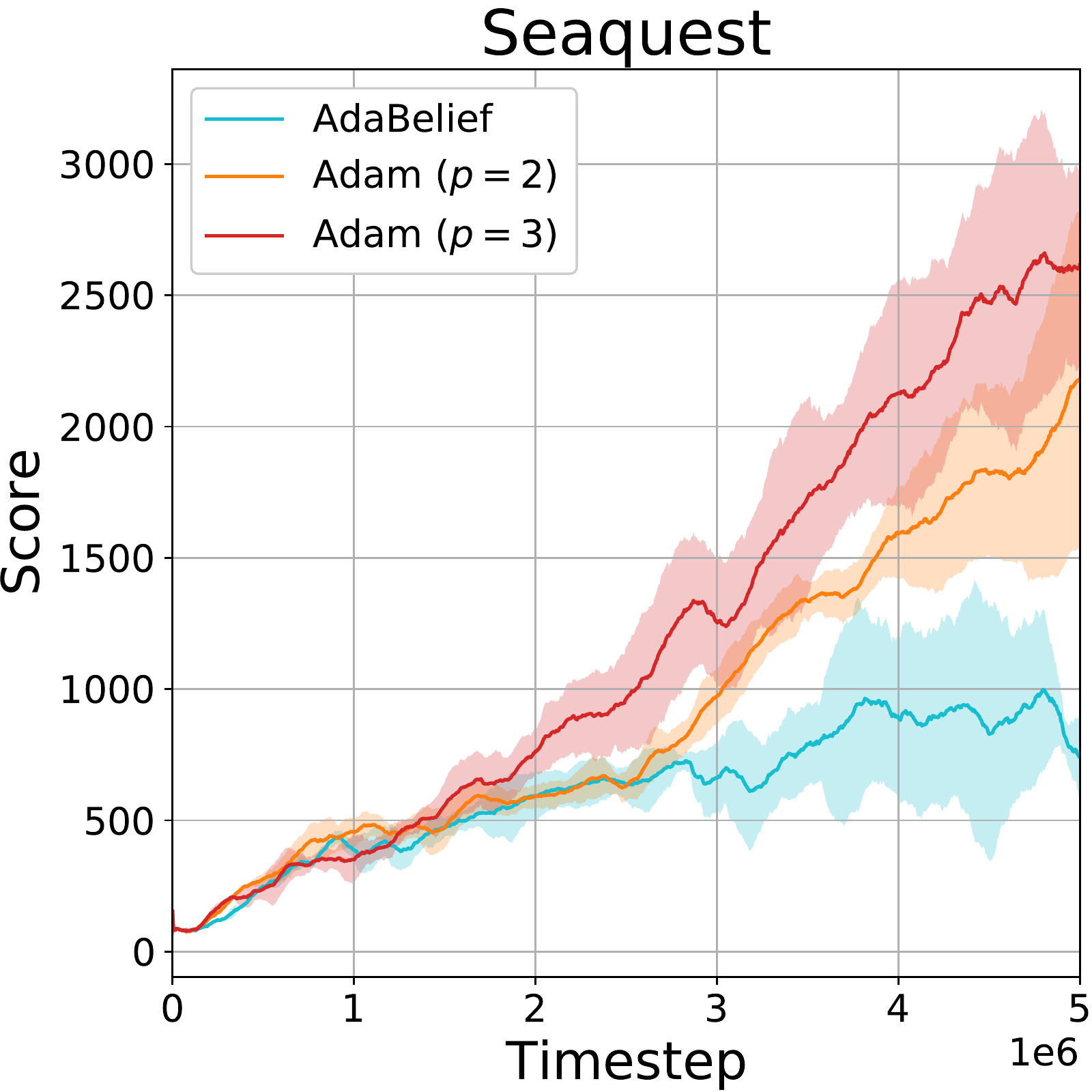}
        \hfill
        \includegraphics[width=0.333\textwidth]{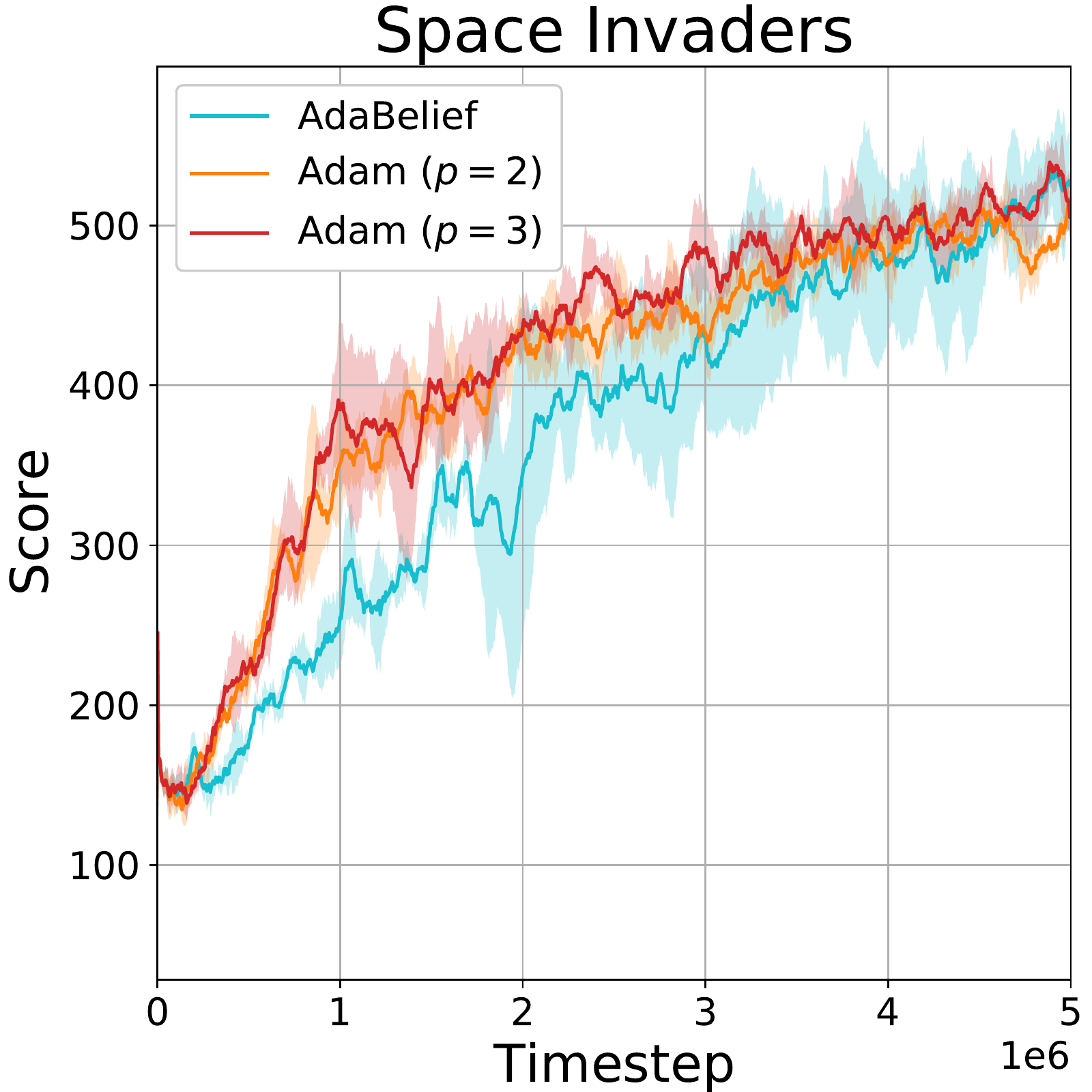}
    }
    \caption{
        Performance comparison of $L^3$ Adam against Adam and AdaBelief when training DQN to play six different Atari 2600 games.
        Results are averaged over 3 trials with standard deviation shaded.
        \label{fig:dqn}
    }
\end{figure}

\tightsection{Experiments}
\label{section:experiments}

Our preceding analysis predicts that changing Adam's $L^p$ norm modifies its step size and direction, but leaves its other properties largely unaffected.
In this section, we conduct a series of deep learning experiments to study how these differences impact empirical performance.
We intentionally choose from a diverse set of problems encompassing supervised, unsupervised, and reinforcement learning in order to vary the encountered optimization challenges as much as possible.
This is crucial, since we are not solely interested in characterizing the behavior of different $L^p$ norms, but would also like to identify at least one that generally outperforms the standard $L^2$ norm.

\tightsubsection{Image classification}
\label{sect:img_class}

We started our classification experiments with the classic MNIST handwriting recognition task~\citep{lecun1998gradient}.
The training set consists of 60,000 grayscale images of size $28 \times 28$ that are labeled according to which digit they contain.
A good classifier for this task can be trained without parameter sharing (such as convolutional layers).
We preprocessed the training data by subtracting the mean image taken across all samples.
We then trained a 2-layer, 1,000-unit fully connected network (labeled ``FC-2'') with rectified linear unit (ReLU) activations, matching the architecture from~\cite{kingma2015adam}.
The network was trained for 100 epochs on randomly sampled minibatches of 128 images.

Since MNIST is relatively easy by modern standards, we also conducted experiments on the CIFAR-10 dataset~\citep{krizhevsky2009learning}.
The training set contains 50,000 colored $32 \times 32$ images that must be classified as one of 10 objects.
We trained two large convolutional neural networks:
a 16-layer VGGNet~\citep{simonyan2014very} with 33.6 million parameters and a 50-layer ResNet~\citep{he2016deep} with 23.5 million parameters.
The juxtaposition of these architectures is interesting because ResNet's residual connections drastically change the internal topology compared to VGGNet's strict sequential structure, and we expect their optimizations to behave differently as a result.
Both networks were trained for 500 epochs on minibatches of 256 images.

We compared Adam's performance using five different $L^p$ norms (see Table~\ref{table:p_values}), where we also included AdaMax as a baseline to represent $p \to \infty$.
We chose the best learning rate for each method based on a logarithmic search (see Appendix~\ref{app:hyperparameters}).
In Figure~\ref{fig:img_class_lp}, we plot the logarithm of the training loss after each epoch, where we observe an interesting trend:
increasing $p$ leads to better final performance on all three tasks, with $p=6$ being the clear best choice, outperforming AdaMax in the CIFAR-10 tasks.
The $L^6$ norm also outperforms four recent state-of-the-art methods in the MNIST task, and performs competitively in the CIFAR-10 tasks (Figure~\ref{fig:img_class_sota}).

\tightsubsection{Language modeling}

We trained 1-, 2-, and 3-layer, 200-unit LSTM models~\citep{hochreiter1997long} inspired by the architecture from~\cite{zaremba2014recurrent} to perform next-word prediction on the Penn Treebank corpus~\citep{marcus1993building}.
The sequential nature of this modeling task makes optimization difficult, as recurrent neural networks are known to be susceptible to vanishing and exploding gradients~\citep{bengio1994learning}.
While the LSTM can partially overcome these difficulties, it also features saturating nonlinearities (sigmoid and tanh) that were notably absent from our image classification experiments.
Stacking multiple LSTM layers compounds these challenges, so it is interesting to see how network depth affects optimization performance.
We trained the networks for 200 epochs on minibatches of 20 sequences of 20 words each, where words were embedded as 200-dimensional vectors.

We conducted experiments analogous to those in Section~\ref{sect:img_class}.
The $L^p$-norm comparison is shown in Figure~\ref{fig:ptb_lp};
interestingly, we do not observe the same trend we saw in the image classification tasks.
Here, all methods including AdaMax perform similarly to or worse than the $L^2$ Adam baseline, with the notable exception of $L^6$ Adam, whose performance gain is even more dramatic than before.
In Figure~\ref{fig:ptb_sota}, we once again compare the $L^6$ norm against state-of-the-art baselines, where it outperforms all of them by a significant margin.

\tightsubsection{Deep reinforcement learning}

For our final experiments, we trained Deep Q-Network (DQN)~\citep{mnih2015human} to play six of the Atari 2600 games:
Beam Rider, Breakout, Pong, Q*Bert, Seaquest, and Space Invaders.
Unlike our previous experiments, these problems are highly nonstationary.
The network is trained on minibatches of experiences stored in a dataset $D$ that has finite capacity, meaning that old samples are overwritten as the agent explores its environment.
Additionally, the agent's actions are dependent on the network parameters $\theta$, and therefore the distribution of new data entering $D$ changes as the network is trained.
Optimization methods must effectively balance convergence with responsiveness to distributional shifts in order to perform well.

We followed the experimental setup described in~\cite{mnih2015human} including preprocessing, reward clipping, exploration, and hyperparameters (except for the optimization).
The network architecture consisted of three convolutional and two fully connected layers with ReLU activations, for a total of nearly 4 million parameters.
We trained the agents for 5 million timesteps (20 million game frames).

In Figure~\ref{fig:dqn}, we compare $L^3$ Adam (which we found to be the most successful for these experiments) against Adam and AdaBound.
AdaBound was shown to outperform Adam in a related deep reinforcement learning setting, making for an interesting side-by-side comparison.
We plot the moving average of the 100 previous episode scores versus completed timesteps, following the recommendation of~\cite{machado2018revisiting}.
In three of the games (Beam Rider, Pong, and Seaquest), $L^3$ Adam resulted in a significant score improvement compared to Adam, and it achieved comparable performance in the remaining three games.
We observed that AdaBound, surprisingly, was not competitive with either of the Adam variants for these problems.

\tightsection{Discussion}

In general, our experiments show that the performance of Adam can be substantially improved by $L^p$ normalization with $p > 2$.
In no instance did a choice of $p < 2$ offer a noticeable benefit in our experiments.
However, these results do not imply that performance improves monotonically with increasing $p$.
For example, in our language modeling experiments, we observed strong performance when using $p=6$, but also observed that $p=3$ was not able to substantially outperform the $p=2$ baseline.
This is because the methods did not always share the same best learning rate, which interacts with the $L^p$ norm to determine the overall step magnitude.

Had we selected the same learning rate for all the $L^p$ norms, we expect that the trend \textit{would} be monotonic in $p$ based on our analysis in Section~\ref{sect:analysis}---at least up until some reasonable, finite value.
We do not expect arbitrarily large values of $p$ to yield increasingly better performance, based on the results we obtained for AdaMax (since this method represents the limiting case as $p \to \infty$).
While AdaMax demonstrated fast initial learning in the classification experiments, it was eventually surpassed by the $L^6$ norm when training large convolutional models on the CIFAR-10 dataset.
Later, in the language modeling tasks where the $L^6$ norm excelled, AdaMax's performance failed to offer any improvement over the Adam baseline.
This leads us to believe that there is a tradeoff between low and high values of $p$, and therefore intermediate values work best in practice.
Future work should explore more $p$-values, particularly with $p > 6$, to identify candidates for even stronger performance.

While $p=6$ worked very well for the supervised and unsupervised learning tasks, we needed to reduce the value to $p=3$ to obtain good performance for the reinforcement learning problems.
This makes sense given that the image classification and language modeling tasks are stationary, whereas reinforcement learning is not.
Smaller steps are beneficial when the underlying data distribution is static, because the optimization method has time to gradually settle into an accurate solution.
In contrast, when the data distribution is changing, larger steps may be necessary simply to keep up with the evolving loss landscape.
Failure to do so is particularly pernicious in the reinforcement setting, where poor policy parameters cause the agent to take suboptimal actions and collect poor data, resulting in a negative feedback loop.
This presumably explains why the $L^6$ norm did not perform as well in this setting.
It is possible that an intermediate value ($3 < p < 6$) could balance performance between these types of tasks simultaneously, although we did not try this.

Ultimately, our experiments demonstrate that a choice of $p \in \{3,6\}$ can significantly improve performance, regardless of architectural details or task nature.
To conclude, we note that the $L^3$ norm has a meaningful interpretation:
steps are bounded along each dimension by exactly the learning rate $\alpha$ (recall Table~\ref{table:p_values}).
In contrast, Adam typically takes steps that are bounded by $\smash{ \sqrt{10} \alpha }$.
Our experiments suggest that this is too aggressive in practice;
furthermore, this cannot be rectified simply by reducing the learning rate, as our hyperparameter sweep revealed that no value enabled Adam to match the performance of the $L^3$ or $L^6$ norms in these tasks.
This offers empirical evidence that the nonlinear effect of $p$ on the magnitude and direction of the steps is crucial for performance.

\tightsection{Conclusion}

We theoretically and empirically analyzed the $L^p$-norm generalization of Adam, which was proposed in~\cite{kingma2015adam} but unexplored until now.
We verified that these $L^p$ variants are well motivated and showed that they preserve important properties of Adam, while differing primarily in how they affect the relative magnitude and direction of the optimization steps.

Our experiments demonstrated that the choice of $L^3$ or $L^6$ improved both learning speed and final performance compared to Adam and AdaMax in a variety of deep learning tasks.
Furthermore, they outperformed state-of-the-art methods.
The $L^6$ norm was evidently the best choice for the supervised and unsupervised tasks we tested, but the $L^3$ norm was better for deep reinforcement learning---most likely because larger steps are necessary to track the nonstationary data distribution.

Finally, we note that while our work focused specifically on Adam, other methods based on RMS normalization could be similarly generalized to utilize an $L^p$-norm update, including all of the baselines we tested in our experiments;
this could be an interesting avenue for future work.
Another promising possibility is a dynamically scheduled $p$-value;
for example, a variant of Adam that benefits from larger steps ($p < 2$) early in training before transitioning to smaller steps ($p > 2$) later.



\bibliographystyle{plainnat}
\small
\bibliography{references}


\clearpage
\appendix
\section{Hyperparameters}
\label{app:hyperparameters}

The methods were tested with five learning rates:
$3 \times 10^{-5}, 10^{-4}, 3 \times 10^{-4}, 10^{-3}, 3 \times 10^{-3}$.
This corresponds to an approximate geometric search over two orders of magnitude, where each learning rate is (roughly) triple the previous.
We reported the one that yielded the best final performance for each method (see Tables~\ref{table:lr_supervised} and~\ref{table:lr_reinforcement} below).
Our preliminary results showed that values outside of this range performed poorly for all of the tested methods, allowing us to focus the search on this interval.

All methods added the same constant $\epsilon = 10^{-8}$ to their denominators for numerical stability.
Any other hyperparameters were left as the default values from the methods' respective papers.

\begin{table}[h]
    \centering
    \caption{Best learning rates for the image classification and language modeling experiments.}
    \begin{tabular}{l *7r}
        \toprule
        Method & FC-2 & VGGNet-16 & ResNet-50 & LSTM (all) \\
        \midrule
        AdaBelief               & $3 \times 10^{-3}$ & $3 \times 10^{-4}$ & $3 \times 10^{-4}$ & $3 \times 10^{-3}$ \\
        AdaBound                & $1 \times 10^{-3}$ & $1 \times 10^{-4}$ & $1 \times 10^{-4}$ & $3 \times 10^{-5}$ \\
        AMSGrad                 & $1 \times 10^{-3}$ & $1 \times 10^{-4}$ & $3 \times 10^{-4}$ & $1 \times 10^{-3}$ \\
        Yogi                    & $3 \times 10^{-3}$ & $3 \times 10^{-4}$ & $3 \times 10^{-4}$ & $3 \times 10^{-3}$ \\
        Adam ($p=1.2$)          & $3 \times 10^{-5}$ & $3 \times 10^{-5}$ & $1 \times 10^{-4}$ & $3 \times 10^{-4}$ \\
        Adam ($p=1.5$)          & $3 \times 10^{-5}$ & $1 \times 10^{-4}$ & $1 \times 10^{-4}$ & $3 \times 10^{-4}$ \\
        Adam ($p=2$)            & $3 \times 10^{-5}$ & $1 \times 10^{-4}$ & $3 \times 10^{-4}$ & $1 \times 10^{-3}$ \\
        Adam ($p=3$)            & $1 \times 10^{-4}$ & $1 \times 10^{-4}$ & $3 \times 10^{-4}$ & $1 \times 10^{-3}$ \\
        Adam ($p=6$)            & $3 \times 10^{-4}$ & $3 \times 10^{-4}$ & $3 \times 10^{-4}$ & $3 \times 10^{-3}$ \\
        AdaMax ($p \to \infty$) & $1 \times 10^{-3}$ & $3 \times 10^{-4}$ & $3 \times 10^{-4}$ & $3 \times 10^{-3}$ \\
        \bottomrule
    \end{tabular}
    \label{table:lr_supervised}
\end{table}

\begin{table}[h]
    \centering
    \caption{Best learning rates for the deep reinforcement learning experiments.}
    \begin{tabular}{l *7r}
        \toprule
        Method & Beam Rider & Breakout & Pong & Q*Bert & Seaquest & Space Invaders \\
        \midrule
        AdaBelief               & $1 \times 10^{-4}$ & $1 \times 10^{-4}$ & $3 \times 10^{-4}$ & $1 \times 10^{-4}$ & $3 \times 10^{-4}$ & $1 \times 10^{-4}$\\
        Adam ($p=2$)            & $3 \times 10^{-5}$ & $1 \times 10^{-4}$ & $1 \times 10^{-4}$ & $1 \times 10^{-4}$ & $3 \times 10^{-5}$ & $1 \times 10^{-4}$\\
        Adam ($p=3$)            & $3 \times 10^{-5}$ & $1 \times 10^{-4}$ & $1 \times 10^{-4}$ & $1 \times 10^{-4}$ & $1 \times 10^{-4}$ & $1 \times 10^{-4}$\\
        \bottomrule
    \end{tabular}
    \label{table:lr_reinforcement}
\end{table}

\end{document}